\documentclass[11pt, a4paper, logo, twocolumn, copyright]{googledeepmind}

\pdftrailerid{redacted}

\makeatletter
\renewcommand\bibentry[1]{\nocite{#1}{\frenchspacing\@nameuse{BR@r@#1\@extra@b@citeb}}}
\makeatother

\usepackage{longtable}
\usepackage{dsfont}
\usepackage{gdm-colors}
\usepackage{multirow}
\usepackage{graphicx}
\usepackage{rotating}
\usepackage{geometry}
\usepackage{xcolor}
\usepackage{placeins}

\usepackage[numbers, sort&compress, square]{natbib}

\makeatletter
\DeclareRobustCommand\onedot{\futurelet\@let@token\@onedot}
\def\@onedot{\ifx\@let@token.\else.\null\fi}

\def\eg/{\emph{e.g}\onedot} \def\Eg/{\emph{E.g}\onedot}
\def\ie/{\emph{i.e}\onedot} \def\Ie/{\emph{I.e}\onedot}
\def\cf/{\emph{c.f}\onedot} \def\Cf/{\emph{C.f}\onedot}
\def\etc/{\emph{etc}\onedot} \def\vs/{\emph{vs}\onedot}
\def\wrt/{w.r.t\onedot} \def\dof/{d.o.f\onedot}
\def\etal/{\emph{et al}\onedot}
\newcommand{\pg}{PaliGemma }
\newcommand{\pgnospace}{PaliGemma}
\newcommand{\pgtwo}{PaliGemma~2 }
\newcommand{\pgtwonospace}{PaliGemma~2}

\newcommand{\modelres}[1]{\textcolor{gray}{\scriptsize{#1px$^2$}}}

\makeatother

\newcolumntype{L}[1]{>{\raggedright\let\newline\\\arraybackslash\hspace{0pt}}m{#1}} 

\title{PaliGemma 2:\\A Family of Versatile VLMs for Transfer}

\newcommand{\titlerunning}{PaliGemma 2: A Family of Versatile VLMs for Transfer}

\correspondingauthor{andstein,andresp,tschannen@google.com}

\renewcommand{\today}{December 2024}

\author[*,$\dagger$]{Andreas Steiner}
\author[*]{André Susano Pinto}
\author[*]{Michael Tschannen}
\author[ \hspace{-0.7ex}]{Daniel Keysers}
\author[ \hspace{-0.7ex}]{Xiao Wang}
\author[ \hspace{-0.7ex}]{Yonatan Bitton}
\author[ \hspace{-0.7ex}]{Alexey Gritsenko}
\author[ \hspace{-0.7ex}]{Matthias Minderer}
\author[ \hspace{-0.7ex}]{Anthony Sherbondy}
\author[ \hspace{-0.7ex}]{Shangbang Long}
\author[ \hspace{-0.7ex}]{Siyang Qin}
\author[ \hspace{-0.7ex}]{Reeve Ingle}\author[ \hspace{-0.7ex}]{Emanuele Bugliarello}
\author[ \hspace{-0.7ex}]{Sahar Kazemzadeh}
\author[ \hspace{-0.7ex}]{Thomas Mesnard}
\author[ \hspace{-0.7ex}]{Ibrahim Alabdulmohsin}
\author[ \hspace{-0.7ex}]{Lucas Beyer}
\author[ \hspace{-0.7ex}]{Xiaohua Zhai}

\affil[ \hspace{-0.7ex}]{Google DeepMind}
\affil[*]{Core team}
\affil[$\dagger$]{Project lead}

\begin{abstract}
\pgtwo is an upgrade of the \pg open Vision-Language Model (VLM) based on the Gemma~2 family of language models.
We combine the SigLIP-So400m vision encoder that was also used by \pg with the whole range of Gemma 2 models, from the 2B one all the way up to the 27B model.
We train these models at three resolutions (224px$^2$, 448px$^2$ and 896px$^2$) in multiple stages to equip them with broad knowledge for transfer via fine-tuning. The resulting family of base models covering different model sizes and resolutions allows us to investigate factors impacting transfer performance (such as learning rate) and to analyze the interplay between the type of task, model size, and resolution.
We further increase the number and breadth of transfer tasks beyond the scope of \pg including different OCR-related tasks such as 
table structure recognition, molecular structure recognition, music score recognition, as well as long fine-grained captioning and radiography report generation, on which \pgtwo obtains state-of-the-art results.
\end{abstract}

\begin{document}

\maketitle

\section{Introduction}

\pgnospace~\cite{beyer2024paligemma} is a 3B vision-language model (VLM) for transfer combining the SigLIP~\cite{siglip} vision encoder and the 2B Gemma language model~\cite{gemma}. It matches the performance of much larger prior VLMs consisting of a range of different vision encoders and language models. We now upgrade \pg by replacing its language model component with the more recent and more capable language models from the Gemma~2 family~\cite{gemmateam2024gemma2},
producing new \pgtwo base VLMs at 3 different sizes (3B, 10B, 28B) and 3 different resolutions (224px$^2$, 448px$^2$, 896px$^2$).
To equip these VLMs with broad capabilities we use the same 3-stage training recipe as \pgnospace.
The resulting models are designed to be fine-tuned, and when evaluated on the 30+ transfer tasks considered in~\cite{beyer2024paligemma} (which include common captioning and VQA tasks, and some video and referring expression tasks), \pgtwo slightly outperforms \pg at the same resolution and model size, and obtains substantial improvements
at larger model sizes. We release the \pgtwo VLMs as open-weight models which can serve as drop-in replacement for \pgnospace.

Having a family of models at hand that are all derived from comparable building blocks and are trained according to the same recipe allows us to analyze the effect of model size and resolution on the downstream performance in a controlled setting (see Sec.~\ref{sec:res_model_size}).
For example, while almost every task benefits from added compute, we identify which transfer tasks benefit more from compute due to increased resolutions, and which from compute due to a larger, more capable language model.
We also show that larger models tend to have a lower optimal \emph{transfer} learning rate.

We also explore new tasks which were not explored in depth in~\cite{beyer2024paligemma}, including text detection and recognition (Sec.~\ref{sec:ocr}), table structure recognition (Sec.~\ref{sec:tables}), molecular structure recognition (Sec.~\ref{sec:molecules}), optical music score recognition (Sec.~\ref{sec:music}), 
long caption generation (Sec.~\ref{sec:docci}), spatial reasoning (Sec.~\ref{sec:spatial}), and radiography report generation (Sec.~\ref{sec:cxr}). \pgtwo obtains state-of-the-art results on many of those tasks. Finally, we benchmark and analyze low-precision variants of \pgtwo for on-device deployment on CPU (Sec.~\ref{sec:cpu_inference}).

\section{Related work}

Over the last few years, VLMs evolved rapidly from simple dual-encoder (contrastive)~\cite{clip, align, siglip} or encoder-decoder (captioning)~\cite{wang2021simvlm, desai2021virtex, cappa, locca} designs trained from scratch, to more capable designs combining a pretrained vision encoder with a pretrained language model~\cite{alayrac2022flamingo, wang2022git, peng2023kosmos, blip2, qwen-vl, pali, palix, ye2024mplug}. Broadly, three paradigms are used to transfer these models: zero-shot, few-shot, and fine-tuning. Another recent trend is ``instruction tuning'' which aims to make the models more user friendly~\cite{llava1, dai2023instructblip}.

Several previous works~\cite{idefics2, mm1, cambrian1, zhang2024mm1, karamcheti2024prismatic, beyer2024paligemma, kar2024brave, deitke2024molmo} have investigated the effect of scaling VLMs along different axes such as training data and compute, resolution, model size, and quality of components, in particular the vision encoder. However, we are not aware of prior work which jointly studies the effect of the image resolution and the size of the language models on \emph{transfer via fine-tuning}. In particular, prior works relying on different language model sizes often use models with different architecture and training recipes from different labs, e.g.~\cite{cambrian1, karamcheti2024prismatic} (with the notable exception of~\cite{li2024llavanext-ablations}).

\vspace{-0.3cm}
\section{Model}

\begin{figure}
  \centering\vspace{-0.2cm}
  \includegraphics[width=\columnwidth]{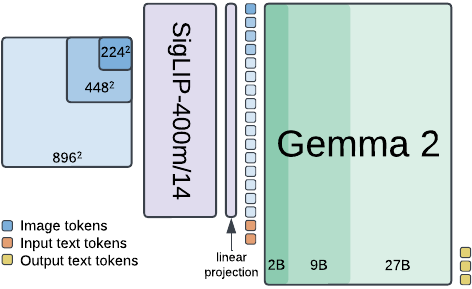}
  \caption{\pgtwo processes a 224px$^2$/ 448px$^2$/896px$^2$ image with a SigLIP-400m encoder with patch size 14px$^2$, yielding 256/1024/ 4096 tokens. After a linear projection, the image tokens are concatenated with the input text tokens and Gemma~2 autoregressively completes this prefix with an answer. \vspace{-0.2cm}}
  \label{fig:model}
\end{figure}

\begin{figure}
  \centering
  \includegraphics[width=\columnwidth]{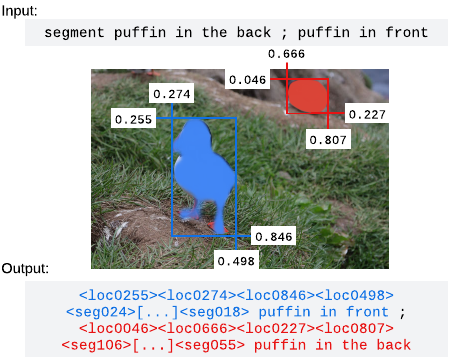}
  \caption{Referring segmentation example from our \pg demo\textsuperscript{a}.
  The model is pretrained with a vocabulary that includes localization tokens (for detection) and segmentation tokens (to define a binary mask inside a bounding box).}
  \small\textsuperscript{a} \resizebox{0.9\columnwidth}{!}{\url{https://huggingface.co/spaces/big-vision/paligemma}}
  \label{fig:segmentation}
\end{figure}

\begin{table*}
\centering
\begin{tabular}{l c c c c c c}
\toprule
 & & & & \multicolumn{3}{c}{Training cost / example} \\
\cmidrule(lr){5-7}
 &  Vision Encoder & LLM & Params. & 224px$^2$ & 448px$^2$ & 896px$^2$ \\
\midrule
\pgtwo~\phantom{1}3B  &               & Gemma 2~\phantom{2}2B  &  \phantom{2}3.0B &  \phantom{1}1.0 &  \phantom{1}4.6 & \phantom{$\sim$1}23.5 \\
\pgtwo~10B & SigLIP-So400m & Gemma 2~\phantom{2}9B  &  \phantom{2}9.7B &  \phantom{1}3.7 & 18.3 & \phantom{$\sim$1}67.7 \\
\pgtwo~28B &               & Gemma 2~27B & 27.7B & 18.9 & 63.5 &      $\sim$155.6\\
 \bottomrule
\end{tabular}
    \caption{The vision encoder parameter count is small compared to the LLM, but the compute is dominated by the vision tokens in the LLM. The last three columns show the relative training cost per example (as measured in our pre-training setup). Models are trained on Cloud TPUv5e~\cite{tpucloud}, except the 28B model at 896px$^2$ is trained on TPUv5p, for which we assume a speed-up of $2.3\times$ per chip.}
    \label{tab:arch}
\end{table*}

We follow exactly the same modeling, training, and data setup as PaliGemma~\cite{beyer2024paligemma} and briefly summarize the most important aspects here. We use the same pretrained SigLIP-So400m vision encoder~\cite{siglip, sovit} and map its (sequence of) embeddings to the Gemma~2 input space with a linear projection. The visual embeddings are combined with a text prompt and fed to the Gemma~2 language model (prefill). Predictions are then obtained by autoregressively sampling from the language model (see Fig.~\ref{fig:model}).

We pretrain \pgtwo in three stages (with stage 0 corresponding to unimodal pretraining of the components, see \cite{siglip} and \cite{gemma}). 
\begin{itemize}
    \item {\bf Stage 1} combines the pretrained SigLIP-So400m and Gemma~2 checkpoints (raw checkpoints, without post-training steps) and trains them jointly on a multimodal task mixture of 1 billion examples designed to enable transferability to a wide range of tasks via fine-tuning. The image resolution is 224px$^2$; no parameters are frozen during this stage.
    \item {\bf Stage 2} first trains for 50 million examples at resolution 448px$^2$ and then for 10 million examples at resolution 896px$^2$. The task mixture has the same components but tasks benefiting from high resolution are upweighted, and the output sequence length is increased (to promote e.g.\ learning of OCR for long sequences of visual text).
    \item {\bf Stage 3} fine-tunes the checkpoints from stage 1 or 2 (depending on the resolution) to the target task. PaliGemma considered a range of academic benchmarks, including some involving multiple images and short videos. We consider the same set of benchmarks here (exploring the same set of hyperparameters from \cite[Sec.~3.2.4]{beyer2024paligemma}). In addition, we also explore new applications involving document-related tasks, long caption generation, and medical image understanding.
\end{itemize}

Following~\cite{gemmateam2024gemma2}, we apply logits soft-capping \cite{bello2016neural} to the attention and output logits in the Gemma~2 component with the same parameters as~\cite{gemmateam2024gemma2} in Stages 1 and 2, but not in Stage 3, as this led to worse results for some transfer tasks. Further, we use the Adam optimizer~\cite{kingma2017adam} with default hyperparameters throughout, and adjust the learning rate based on the model size in Stages 1 and 2. Specifically, we multiply the learning rate of $2\cdot10^{-5}$ used in Stages 1 and 2 for \pg by 0.5 for \pgtwo 3B and by 0.25 for \pgtwo 10B and 28B. 

For details on the training data mixture we refer to \cite[Sec.~3.2.5]{beyer2024paligemma} and provide a brief summary here. The mixture involves captioning, grounded captioning (as in~\cite{locca}), OCR, different machine generated visual question answering (VQA) tasks~\cite{Changpinyo, Piergiovanni}, detection~\cite{pix2seq} and instance segmentation~\cite{pali3}. Many of the corresponding labels are machine generated, mostly relying on publicly available specialist models (see \cite[Sec.~3.2.5]{beyer2024paligemma}), and none uses a large commercial VLM as common among other open VLMs such as LLaVA~\cite{llava1}. 

Similar to PaliGemma, we train \pgtwo models on Cloud TPUv5e Pod slices~\cite{tpucloud} (except TPUv5p for the 28B model at 896px$^2$) of 256 to 1024 chips and use a fully-sharded data-parallel (FSDP~\cite{fsdp, big_vision}) sharding strategy. PaliGemma~2 3B has roughly the same training cost as PaliGemma (3 days for Stage 1 using 256 chips); the cost for other variants and resolutions can be inferred from Table~\ref{tab:arch}. It is worth noting that increasing resolution incurs a similar additional cost as increasing the language model size.

\section{Experiments}

\newcommand{\rgbline}[1]{\raisebox{.5ex}{\textcolor[HTML]{#1}{\rule{0.3cm}{1pt}}}}
\begin{figure*}[t]
  \centering
  \includegraphics[width=1.7\columnwidth]{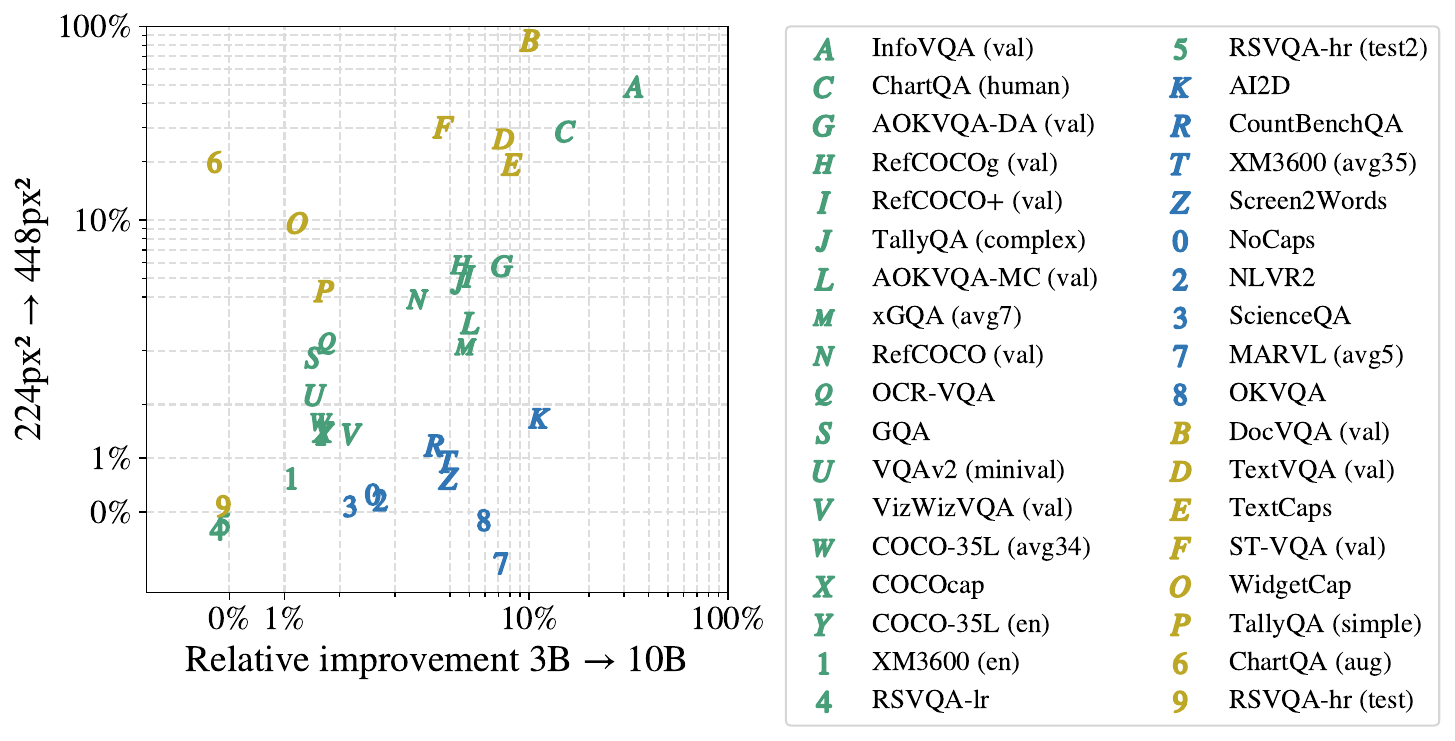}
  \caption{Relative improvements of metrics after transfer, when choosing a pre-trained checkpoint with a larger LM, or with a higher resolution. The tasks are grouped into tasks sensitive to both model size and resolution (\rgbline{479E78}), sensitive to model size (\rgbline{3176B4}), and sensitive to resolution (\rgbline{BDA727}).
  Note that some benchmarks are quite saturated (e.g. ScienceQA's relative improvement of 2.2\% corresponds to an error reduction of 53.8\% -- see Figure~\ref{fig:app:relimp}).
  Data used to create this plot available in Table~\ref{tab:app:pg-results}.
  }
  \label{fig:res:relimp}
\end{figure*}

In addition to the broad range of transfer tasks considered in \cite{beyer2024paligemma}, we also consider new tasks involving text detection and recognition (Sec.~\ref{sec:ocr}), table structure recognition (Sec.~\ref{sec:tables}), molecular structure recognition (Sec.~\ref{sec:molecules}), optical music score recognition (Sec.~\ref{sec:music}), 
long caption generation (Sec.~\ref{sec:docci}), spatial reasoning (Sec.~\ref{sec:spatial}), and radiography report generation (Sec.~\ref{sec:cxr}).

We provide examples for each new task in Appendix~\ref{sec:app_tasks} and transfer details in Appendix~\ref{sec:app_transfer_details}.

\begin{figure*}
  \centering
  \includegraphics[width=\textwidth]{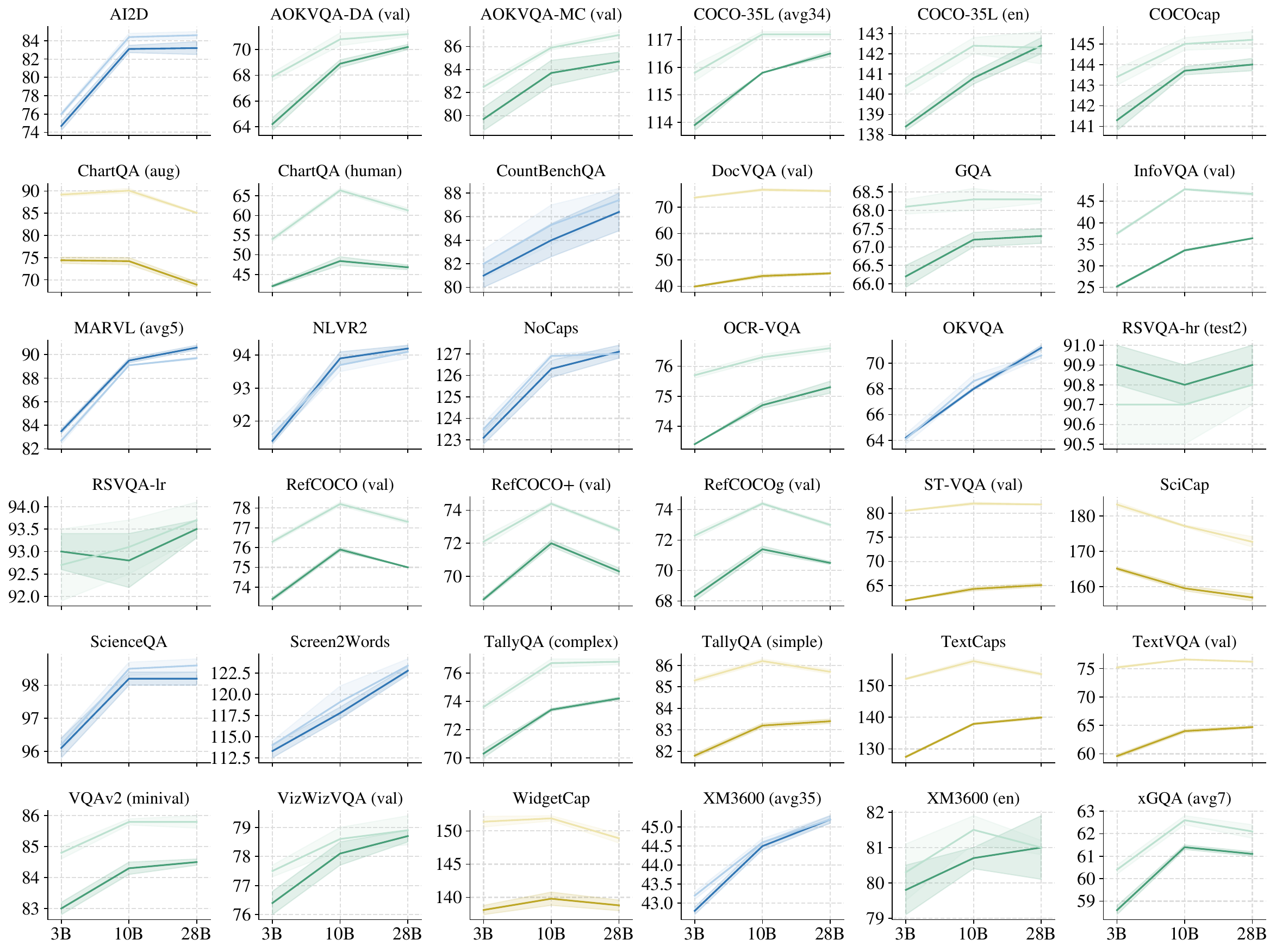}
  \caption{Transfer performance as a function of model size and resolution (median over 5 transfer runs). The shaded area marks standard deviation to reported value. Lighter lines correspond to higher resolution (448px$^2$). The tasks are grouped into tasks sensitive to both model size and resolution (\rgbline{479E78}), sensitive to model size (\rgbline{3176B4}), and sensitive to resolution (\rgbline{BDA727}). Data for this plot is available in Table~\ref{tab:app:pg-results}.}
  \label{fig:res:gridplot}
\end{figure*}

\subsection{Investigating model size and resolution} \label{sec:res_model_size}
To study the effect of model size and resolution on task performance we finetune the 3 model variants (3B, 10B and 28B) in two resolutions (224px$^2$ and 448px$^2$) on the 30+ academic benchmarks used by~\cite{beyer2024paligemma}, covering a broad range of captioning, VQA, and referring segmentation tasks on natural images, documents, infographics, and videos.
We reuse the optimal hyperparameters from the earlier \pg work and only sweep the learning rate 
$\{0.03, 0.06, 0.1, 0.3, 0.6, 1.0, 3.0\}\cdot10^{-5}$
for every model size. Since for most tasks the earlier work used the same hyperparameters for 224px$^2$ and 448px$^2$, we only sweep at 224px$^2$ resolution and reuse the selection for both resolutions. We select the best learning rate based on the respective validation split for each model size and task, then retrain the models and report the test metrics.
Complete results are available in Table~\ref{tab:app:pg-results}.

\subsubsection{Effect on task performance}

Increasing image resolution and increasing LM size both lead to an increase in the FLOPs spent on the prediction (and training, see Table~\ref{tab:arch}) of our \pgtwo{} models. Thus, we generally expect most tasks to benefit from both these changes. On the other hand, some tasks might benefit from more detail in the input (higher resolution) or better language understanding and increased world knowledge provided by a larger LM. To get a more fine-grained understanding of these aspects we visualize in Fig.~\ref{fig:res:relimp} the relative improvement in transfer metrics when equipping \pgtwo{} 3B (224px$^2$) with either the bigger 9B LM while keeping the resolution ($3.7\times$ more FLOPs), or keeping the model size but increasing the resolution to 448px$^2$ ($4.6\times$ more FLOPs).

As expected, most tasks similarly benefit from a resolution and model increase (green markers). There is a group of tasks (yellow markers) focused on text, document, screen and chart understanding which mainly benefit from a resolution increase. The images in the corresponding benchmarks often have a native resolution significantly larger than 224px$^2$, which is aligned with this observation. Another group of tasks (blue markers) mostly benefits from LM size increase. Some of these tasks involve multilingual data (XM3600 (avg35)), or require advanced visual reasoning (AI2D, CountBenchQA, NLVR2).

Fig.~\ref{fig:res:gridplot} provides additional detail on the scaling behavior as a function of resolution and model size. Compared to increasing model size from 3B to 10B, increasing it further to 28B often only leads to moderate improvements, or no improvements at all. Using the largest \pgtwo can thus be useful if one wants to get the best possible performance and has no compute or latency constraints. A possible factor related to the relatively worse transferability of \pgtwo 28B is that the underlying Gemma~2 27B model is trained from scratch, as opposed to the 2B and 9B models, which are distilled~\cite[Sec.~6.1]{gemmateam2024gemma2}.

\subsubsection{Model size and transfer learning rate} \label{sec:transfer_lr}
\begin{figure*}[t]
  \centering
  \includegraphics[width=\textwidth]{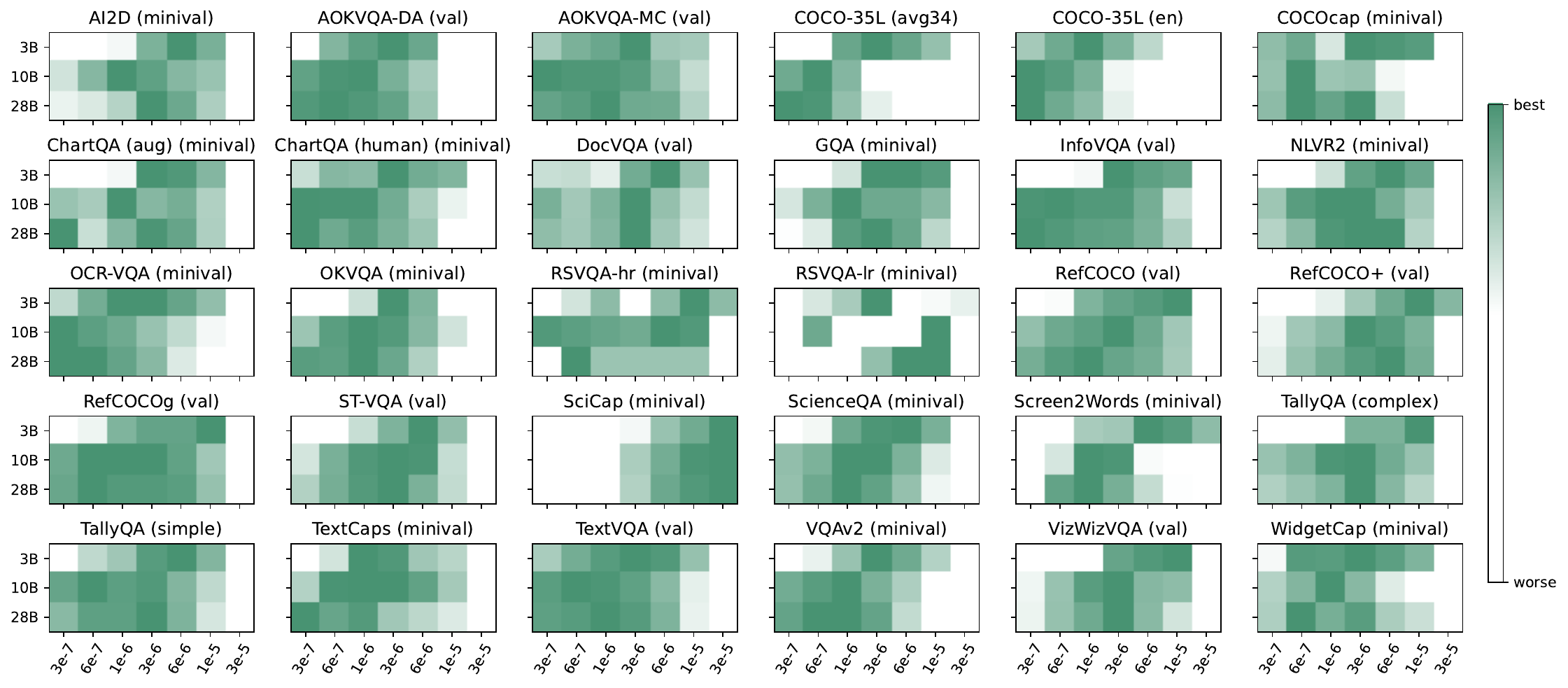}
  \caption{Per-task performance as a function of model size and learning rate for several of the downstream tasks. Values are normalized for each task and model size, with darker color indicating better task performance. Larger models tend to have a lower optimal transfer learning rate. Zero-shot tasks not shown as their values were not used to select learning rates. The data used for this plot is provided in Table~\ref{tab:app:pg_lr_sweep}.}
  \label{fig:lr-sweep}
\end{figure*}
Figure~\ref{fig:lr-sweep} visualizes the (normalized) task performance as a function of the transfer learning rate. As a general trend we observe that the optimal learning rate for larger models tends to be lower than for smaller models (diagonal patterns in the heat map). We thus recommend to sweep smaller learning rates when increasing model size. Additionally, we found that the new \pgtwo 3B generally has a smaller optimal transfer learning rate when compared to \pgnospace.

\subsubsection{Using Gemma~2 instead of Gemma~1}
We also compare with \pg in Table~\ref{tab:app:pg1_pg2}. It can be seen that for the same resolution and model size (i.e. 3B) \pgtwo models perform slightly better than the corresponding \pg models. On average over the 30+ academic benchmarks the scores were 0.65 better for 224px$^2$ and 0.85 for 448px$^2$.

\subsection{Text detection and recognition} \label{sec:ocr}

We apply \pgtwo to advanced OCR involving localization and recognition of individual words from images. Specifically, the outputs are pairs of \{transcription, bounding box\}. Following the HierText competition~\cite{long2023icdar}, we use word level precision, recall, and F1 as the metrics. A word result is considered true positive if the IoU with the ground-truth bounding box is greater than or equal to 0.5 and the transcription matches the ground-truth. Note that the HierText protocol does not normalize letter cases, punctuation symbols, or filter based on text lengths but directly compares predictions against ground-truth.

We fine-tune \pgtwo on a mixture of the train splits of ICDAR'15~\cite{karatzas2015icdar}, Total-Text~\cite{ch2017total},
MLT17 and MLT19~\cite{nayef2017icdar2017}, HierText~\cite{long2022towards}, TextOCR~\cite{singh2021textocr}, IntelOCR~\cite{krylov2021open} and evaluate on the ICDAR'15 and Total-Text test sets, which are the most commonly used OCR benchmarks. Table~\ref{tab:res:ocr} shows the results: \pgtwonospace{}~3B at 896px$^2$ outperforms the state of the art HTS~\cite{long2024hts}. We emphasize that this result is obtained simply by fine-tuning a general-purpose VLM which does not rely on task-specific architecture components as common in the OCR literature. This highlights \pgtwonospace's versatile interface, and shows the benefits of OCR-related pretraining in Stages 2 and 3.  We further tried reducing the resolution which led to substantially lower prediction quality, while increasing the model size did not lead to improvements.

\begin{table*}[t]
  \centering
\begin{tabular}{lcccccc}
\toprule
 & \multicolumn{3}{c}{ICDAR'15 Incidental} & \multicolumn{3}{c}{Total-Text} \\
 \cmidrule(lr{0.5ex}){2-4}
 \cmidrule(lr{0.5ex}){5-7}
 & P & R & F1 & P & R & F1 \\
\midrule
HTS & 81.9 & 68.4 & 74.5 & 75.7 & 69.4 & 72.4 \\
PaliGemma 2 3B \modelres{896} & 81.9 & 70.7 & 75.9 & 73.8 & 74.5 & 74.2 \\
\bottomrule
\end{tabular}
  \caption{Text detection and recognition performance: The 896px$^2$ \pgtwo model outperforms the state-of-the-art model HTS~\cite{long2024hts} on ICDAR'15 Incidental and Total-Text, under the evaluation protocol of HierText~\cite{long2023icdar}.}
  \label{tab:res:ocr}
\end{table*}

\begin{table*}[t]
  \centering
{\begin{tabular}{lcccccccc}
\toprule
 & \multicolumn{4}{c}{FinTabNet} & \multicolumn{4}{c}{PubTabNet} \\
 \cmidrule(lr{1ex}){2-5}  \cmidrule(lr{1ex}){6-9}
 & \footnotesize S-TEDS & \footnotesize TEDS & \footnotesize GriTS-Top & \footnotesize GriTS-Con & \footnotesize S-TEDS & \footnotesize TEDS & \footnotesize GriTS-Top & \footnotesize GriTS-Con \\
\midrule
SOTA & 98.9 & 98.2 & 99.0 & 98.6 & 97.9 & 96.9 & - & - \\
PaliGemma 2 3B \modelres{896} & 99.2 & 98.9 & 99.4 & 99.2 & 97.6 & 97.3 & 98.0 & 97.8 \\
\bottomrule
\end{tabular}}
  \caption{\pgtwo results for table structure recognition on FinTabNet~\cite{fintabnet} and PubTabNet~\cite{teds}, compared to the state of the art. The reference metrics are from \cite{huang2023improving, smock2023aligning, ly2023end, kawakatsu2024multi}. }
  \label{tab:res:table_recognition}
\end{table*}

\subsection{Table structure recognition} \label{sec:tables}
The goal of table structure recognition is to extract table text content, corresponding bounding box coordinates, and the table structure in HTML format from document images. To transfer \pgtwo to this task we finetune on (the train splits of) two popular data sets, PubTabNet~\cite{teds} containing 516k images of tabular data from the PubMed Central Open Access Subset (commercial use collection) and FinTabNet~\cite{fintabnet}, consisting of 113k financial report tables from annual reports of S\&P 500 companies. We remove examples with obviously corrupted ground truth (e.g. a bounding box extending outside the image frame) from the training data and further apply the refinements from~\cite{smock2023aligning} to FinTabNet. Images are resized to the target resolution while preserving the aspect ratio, and padded to square size to match the target input resolution.

We assess model quality with the Tree Edit Distance Similarity (TEDS)~\cite{teds} and the Grid Table Similarity (GriTS)~\cite{grits}, two families of metrics which measure cell text content, cell topology/structure, and bounding box quality. \pgtwo sets a new state of the art for most of these metrics (Table~\ref{tab:res:table_recognition}). We further tried increasing the model size which did not lead to additional benefits, and using a lower image resolution led to a small regression in quality.

\vspace{-0.5cm}\subsection{Molecular structure recognition} \label{sec:molecules}

We explore \pgtwo for molecular structure recognition, the task of inferring the molecule graph structure (represented as a SMILES string~\cite{weininger1988smiles}) from molecular drawings. As training data we use 1 million molecules from the PubChem dataset~\cite{kim2016pubchem}, rendered using the Indigo toolkit~\cite{pavlov2011indigo}, and augmented with a variety of drawing styles and random perturbations, following MolScribe~\cite{qian2023molscribe}. We then evaluate on the same eval set as~\cite{qian2023molscribe} consisting of 5.7k synthetic molecule images rendered with the ChemDraw library. We use exact match percentage as a metric, shown in Table~\ref{tab:res:molecules}. \pgtwo outperforms the state of the art MolScribe when using 448px$
^2$ resolution; further increasing the resolution did not lead to a higher exact match percentage. 
\vspace{-0.5cm}

\subsection{Optical music score recognition}  \label{sec:music}

We apply \pgtwo to optical music score recognition: translating images of single-line pianoform scores into their digital score representation in the \texttt{**kern} format\footnote{\url{https://www.humdrum.org/rep/kern/}}. The \texttt{**kern} representation encodes pitch and duration along with other common score-related information such as articulation and barlines.

We use the GrandStaff dataset~\cite{rios2023end} containing 53.7k images and employ the official train, validation and test splits. During training we use both the original images and synthetically augmented versions. Evaluation is done on the original images without distortion. The metrics are the same as in~\cite{sheet_music_transformer} and are based on the the normalized mean edit distance. More specifically, the Character Error Rate (CER) counts errors at the character level, the Symbol Error Rate (SER) measures errors at the symbol level (combining multiple characters), and the Line Error Rate (LER) is based on full lines in the \texttt{**kern} encoding.

The results are shown in Table~\ref{tab:res:music_score} along with those of the current state of the art method~\cite{sheet_music_transformer}. The error rates decrease with increasing resolution, with the best error rates obtained at 896px$^2$ resolution. Increasing the model size from 3B to 10B did not lead to further error reduction.

\subsection{Generating long, fine-grained captions}  \label{sec:docci}
\begin{table}
  \centering
\begin{tabular}{lc}
\toprule
 & Full Match$\uparrow$ \\
\midrule
MolScribe~\cite{qian2023molscribe} & 93.8 \\
PaliGemma 2           10B \modelres{448} & 94.8 \\
\bottomrule
\end{tabular}
  \caption{\pgtwo performance for molecule structure recognition on ChemDraw data~\cite{qian2023molscribe}.}
  \label{tab:res:molecules}
\end{table}

\begin{table}
  \centering
  
\begin{tabular}{lcccccccc}
\toprule
 & CER$\downarrow$ & SER$\downarrow$  & LER$\downarrow$ \\
\midrule
Sheet Music Tr.~\cite{sheet_music_transformer} & 3.9 & 5.1 & 13.1\\
PaliGemma 2 3B \modelres{896}   & 1.6 & 2.3 & \phantom{1}6.7 \\
\bottomrule
\end{tabular}
  \caption{\pgtwo performance for music score recognition on the GrandStaff data set~\cite{sheet_music_transformer}. Character Error Rate (CER), Symbol Error Rate (SER), and Line Error Rate (LER) in [\%].}
  \label{tab:res:music_score}
\end{table}

Generating long image captions with fine-grained detail has many use cases in multimodal learning, for example to train text-to-image generation models with good controllability~\cite{yuscaling, betker2023improving}. To adapt \pg2 for this task we fine-tune on the DOCCI (Descriptions of Connected and Contrasting Images)~\cite{OnoeDocci2024} data set which contains 15k images with detailed human-annotated English descriptions with an average length of 7.1 sentences (639 characters, 136 words). The descriptions provide object spatial relations, object counting, text rendering, world knowledge, etc.

We first fine-tune \pgtwo on DOCCI’s train split, exploring the hyperparameter range suggested in~\cite[Sec.~3.2.4]{beyer2024paligemma}. We select the most performant models by perplexity scores based on the test split, and generate image captions on the 100-image qual\_dev split, with a maximum decoding length of 192. We then conduct human evaluations assessing whether each generated sentence is factually aligned with (entailed by) the image content (see Appendix~\ref{sec:app:docci} for details on the evaluation protocol). Based on these evaluations we select the most factually aligned models and retrain them on the union of train and test splits, followed by another round of human evaluation (on the qual\_dev split). The results, shown in Table~\ref{tab:res:docci} indicate that the fine-tuned \pgtwo model produces more factually aligned sentences than many popular VLMs, which are often instruction-tuned on $10 - 100\times$ larger high-quality captioning sets than \pgtwonospace. Unsurprisingly, we observe that increasing model size and resolution both improve factual alignment.

\begin{table}
  \centering
\setlength{\tabcolsep}{2pt}
\begin{tabular}{lcccc}

\toprule
 & \#par. & \#char. & \#sent. & NES$\downarrow$ \\
\midrule
MiniGPT-4 & \phantom{1}7B & \phantom{1}484 & \phantom{1}5.6 & 52.3 \\
mPLUG-Owl2 & \phantom{1}8B & \phantom{1}459 & \phantom{1}4.4 & 48.4 \\
InstructBLIP & \phantom{1}7B & \phantom{1}510 & \phantom{1}4.0 & 42.6 \\
LLaVA-1.5  & \phantom{1}7B & \phantom{1}395 & \phantom{1}4.2 & 40.6 \\
VILA & \phantom{1}7B & \phantom{1}871 & \phantom{1}8.6 & 28.6 \\
\midrule
PaliGemma & \phantom{1}3B & \phantom{1}535 & \phantom{1}8.9 & 34.3 \\
PaLI-5B & \phantom{1}5B & 1065 & 11.3 & 32.9 \\
\midrule
PaliGemma 2 \modelres{448} & \phantom{1}3B & \phantom{1}529 & \phantom{1}7.7 & 28.4 \\
PaliGemma 2 \modelres{448} & 10B & \phantom{1}521 & \phantom{1}7.5 & 20.3 \\
\bottomrule
\end{tabular}
  \caption{
  \pgtwo results for long captioning on the DOCCI data~\cite{OnoeDocci2024}. Pali* models are models fine-tuned on DOCCI at 448px$^2$; the other baselines are instruction-tuned on a broad range of tasks.
  Average prediction length in characters and sentences, and percentage of Non-Entailment Sentences (NES), measuring factual inaccuracies.\vspace{-\baselineskip}}
  \label{tab:res:docci}
\end{table}

\subsection{Spatial reasoning} \label{sec:spatial}

VLMs like \pg2{} obtain strong performance in vision-language tasks which involve object localization, such as referring expression comprehension and segmentation~\cite{pali3, you2024ferret, locca, beyer2024paligemma}. These tasks and the associated benchmarks often rely on machine-generated annotations and are blind to complex failure modes, e.g.\ those involving negations.

The Visual Spatial Reasoning (VSR) benchmark~\cite{VSR} is designed to overcome these issues and we use it here to assess the spatial reasoning capabilities of \pgtwonospace. It is formulated as a classification task, where a model needs to determine whether a statement about the spatial relationship of objects in the image is correct or not. To use \pgtwonospace's flexible text interface we frame this benchmark as a QA task with True / False answers. The results in Table~\ref{tab:res:localization} show that \pgtwo outperforms prior fine-tuned models, and fine-tuning also provides a significant improvement over InstructBlip~\cite{dai2023instructblip}, a strong zero-shot model form the literature. We observe significant benefits from larger model size, indicating benefits from improved language understanding, whereas going beyond resolution 224 did not lead to improvements.

\begin{table}[t]
  \centering

\begin{tabular}{lcc}
\toprule
& zs. split & rand. split \\
\midrule
Human~\cite{VSR} & \multicolumn{2}{c}{95.4} \\
\midrule
InstructBLIP (zs.)~\cite{dai2023instructblip} & 65.6 & - \\
LXMERT~\cite{tan2019lxmert} & 70.1 & 61.2 \\
\midrule
PaliGemma 2 \phantom{1}3B \modelres{224} & 74.8 & 81.6 \\
PaliGemma 2 10B \modelres{224} & 79.8 & 86.8 \\
\bottomrule
\end{tabular}

  \caption{\pgtwo accuracy on VSR~\cite{VSR} on the zeroshot and random test splits. We show a fine-tuned (LXMERT) and zero-shot (InstructBLIP) baseline from the literature.}
  \label{tab:res:localization}
\end{table}

\subsection{Radiography report generation}  \label{sec:cxr}

To explore the capabilities of \pgtwo models in the medical domain, we apply it to automatic chest X-ray report generation, which can be cast as a (long) captioning task on X-ray images. We fine-tune \pgtwo on the MIMIC-CXR dataset~\cite{johnson2019mimic, goldberger2000physiobank} which contains 377k images (originating from 228k radiographic studies at the Beth Israel Deaconess Medical Center in Boston, MA) with free-text radiology reports. We use the same train, validation, and test splits as \cite{flamingo-cxr}. To improve quality, we use an LLM (Gemini 1.5 pro) to remove mentions of prior X-rays as the model does not have access to those.

We measure the RadGraph F1-score~\cite{jain1radgraph}, which is the F1 score between the entities extracted from the reference report and the generated one using RadGraph. RadGraph takes into account the absence or presence of findings in the report, as well as their relationships to image features. Results are reported on test data held out during training and tuning.

Table~\ref{tab:cxr} shows the performance of \pgtwo models along with baselines from the literature. \pgtwo obtains a state-of-the-art Rad-Graph score. Increasing resolution and model size both lead to modest improvements.

\subsection{CPU inference and quantization} \label{sec:cpu_inference}

\begin{table}[t]
    \centering
    \setlength{\tabcolsep}{3pt}
\begin{tabular}{lcccc}
\toprule
& C$\uparrow$ & B$\uparrow$ & R$\uparrow$ & F1$\uparrow$ \\
\midrule
Flamingo-CXR~\cite{flamingo-cxr} & 13.8 & 10.1 & 29.7 & 20.5 \\
Med-Gemini-2D~\cite{gemini-cxr} & 17.5 & 20.5 & 28.3 & 24.4 \\
\midrule
PaliGemma 2 \phantom{1}3B \modelres{896} & 19.9 & 14.6 & 31.9 & 28.8 \\
PaliGemma 2 10B \modelres{896} & 17.4 & 15.0 & 32.4 & 29.5 \\
\bottomrule
\end{tabular}
    \caption{\pgtwo performance for radiography report generation on the on the MIMIC-CXR data~\cite{johnson2019mimic, goldberger2000physiobank}. We report CIDEr (C), BlEU4 (B), Rouge-L (R), and RadGraph F1-scores [\%]~\cite{jain1radgraph} (a clinical metric).}
    \label{tab:cxr}
\end{table}

In some cases we may want to run inference of \pgtwo on devices without accelerators. 
We are interested in the resulting runtimes and quality when running inference on CPUs, and briefly present experiments using the \texttt{gemma.cpp}\footnote{\url{https://github.com/google/gemma.cpp}} framework here. 
\texttt{gemma.cpp} is a lightweight, portable C++ inference engine that supports 8-bit switched-floating-point  quantization (alternative options for CPU inference include \texttt{llama.cpp}\footnote{\url{https://github.com/ggerganov/llama.cpp}}, XNNPack\footnote{\url{https://github.com/google/XNNPACK}}, and others).

\begin{table*}
  \centering
  \newcommand{\pz}{\phantom{0}}
\begin{tabular}{lcccccc}
\toprule
& & \multicolumn{3}{c}{{Walltime [s]}} & \multicolumn{2}{c}{{Tokens/sec}}  \\
\cmidrule(lr){3-5} \cmidrule(lr){6-7}
{Processor} & {Threads}   & {ViT} & {Prefill} & {Extend} & {Prefill} & {Extend}  \\ 
\midrule
Apple M1 Max & \pz{}4+1 & 1.6\pz & 8.2 & 0.9\pz & \pz{}32 & 12 \\
Apple M3 Pro & \pz{}7+1 & 0.8\pz & 4.4 & 0.5\pz & \pz{}59 & 22\\
AMD Milan   &  \pz{}8+1  & 0.82 & 4.9 & 0.64 & \pz{}53 & 17 \\
AMD Milan   & 32+1  & 0.39 & 1.8 & 0.34 & 144 & 32 \\
AMD Genoa & \pz{}8+1 & 0.36 & 1.8 & 0.29 & 147 & 37  \\
AMD Genoa & 32+1 & 0.17 & 0.8 & 0.27 & 323 & 41  \\
\bottomrule
\end{tabular}
\vspace{-0.5\baselineskip}
  \caption{CPU-only inference speed measurements with \texttt{gemma.cpp}-based implementation on different architectures. Inference of finetuned \pgtwonospace~3B  (224px$^2$) with greedy decoding. Prefill is done with 260 tokens and followed by 11 calls to extend during decoding.}
  \label{tab:res:cpp_speed}
\end{table*}

To assess the inference speed for CPU-only inference, we run \pg2 inference on four different architectures with \texttt{gemma.cpp}.
We use a checkpoint of \pg2 3B (224px$^2$) finetuned on COCOcap and the example image for \pg in \texttt{gemma.cpp}. 
The prompt ``describe this image'' results in a prefill length of $256 + 4 = 260$ tokens (for image + text). 
The output response ``A large building with two towers on the water'' consists of 11 tokens. All runs used batch size 1. 
The results are presented in Table~\ref{tab:res:cpp_speed} and give an overview of what can be expected on different processors (for this particular setting).

\newcommand{\explanation}{Noticeable differences to Table~\ref{tab:app:pg-results} for the Jax version are the result of using greedy decoding for COCOcap and TextCaps.}

\begin{table*}
  \centering
  \begin{tabular}{lccccc}
\toprule
 & COCOcap & TextCaps & AI2D & OKVQA & DocVQA(val) \\
\midrule
Jax, F32, 12.1GB &  140.0      & 126.3       & \phantom{1}75.4      %
& \phantom{1}64.0       & 39.8 \\
\texttt{gemma.cpp}, quantized, 4.0GB &  139.8 & 126.6 & \phantom{1}75.6 & \phantom{1}64.1 & 39.8\\
\midrule
relative metric values [\%]    & \phantom{1}99.9 & 100.2 & 100.1 & 100.1 & 99.9\\
\bottomrule
\end{tabular}
\vspace{-0.5\baselineskip}
  \caption{Quality comparison between Jax/f32 inference on TPU and quantized \texttt{gemma.cpp}-based inference on CPU. Inference of one fine-tuned \pgtwonospace~3B  (224px$^2$) run. \explanation{} Relative numbers based on metric values before rounding to one decimal.}
  \label{tab:res:cpp_quality}
\end{table*}

From evaluations on \pg\cite{beyer2024paligemma} we already know that going from 32-bit floating point (f32) to 16-bit (bf16) weights is possible without a loss of quality.
Here we compare to the \texttt{gemma.cpp} mixed quantization.
Table~\ref{tab:res:cpp_quality} shows a quality comparison for five of the fine-tuning datasets (chosen for coverage of various tasks). 
We fine-tuned \pg2 3B (224px$^2$) once for each of these five datasets. (\explanation) 
We then evaluated the resulting checkpoints both in Jax and in \texttt{gemma.cpp} after quantization. The relative quality after quantization shows no practical quality difference.
\section{Conclusion}

With \pgtwo we present a new family of open-weight models spanning a broad range of model sizes an input resolutions. \pgtwo obtains strong transfer performance across a broad range of captioning, VQA, and video tasks. In particular, the newly added larger variants lead to significant improvements compared to \pg for users with a larger compute budget. Furthermore, we show that \pgtwo excels in applications beyond what was considered in \pgnospace, including domains like music, molecules, and medical imaging.

\FloatBarrier

\bibliographystyle{abbrvnat}
\nobibliography*
\bibliography{document}

\clearpage

\appendix

\newcommand{\subsubcontrib}[1]{\subsection*{#1}\vspace{-5pt}}

{\small
\section*{Contributions and Acknowledgments}

\subsection*{Model development contributors}

\subsubcontrib{Core Contributors}

Andreas Steiner \\
André Susano Pinto \\
Michael Tschannen

\subsubcontrib{Contributors}
Daniel Keysers \\
Xiao Wang \\
Yonatan Bitton \\
Alexey Gritsenko \\
Matthias Minderer \\
Anthony Sherbondy \\
Shangbang Long \\
Siyang Qin \\
Reeve Ingle \\
Emanuele Bugliarello \\
Sahar Kazemzadeh \\
Thomas Mesnard \\
Ibrahim Alabdulmohsin \\
Lucas Beyer \\
Xiaohua Zhai

\subsubcontrib{Lead}
Andreas Steiner

\subsubcontrib{Acknowledgments}
Jan Wassenberg \\
Basil Mustafa

\subsection*{Model release contributors \\and general support}

\subsubcontrib{Gemma Model}
Tris Warkentin \\
Alek Andreev \\
Armand Joulin \\
Victor Cotruta \\
Sanah Choudhry \\
Nathan Byrd

\subsubcontrib{Open Models Success}
Luiz Gustavo Martins \\
Kat Black \\
Phil Culliton \\
Chris Perry \\
D. Sculley \\
Sara Smoot

\subsubcontrib{Marketing}
Glenn Cameron \\
Natalie Dao

\subsubcontrib{Kaggle}
D. Sculley \\
Nilay Chauhan \\
Brenda Flynn \\
Kinjal Parekh

\subsubcontrib{Developer Relations}
Jetha Chan \\
Joe Fernandez \\
Ju-yeong Ji

\subsubcontrib{Keras}
Divyashree Sreepathihalli \\
Hongyu Chiu

\subsubcontrib{Vertex AI}
Keelin McDonell

\subsubcontrib{Ethics and Safety}
Antonia Paterson \\
Pankil Botadra

\subsubcontrib{Hugging Face Partners}
Merve Noyan \\
Pedro Cuenca \\
Pablo Montalvo

\subsubcontrib{Nvidia Partners}
Dong Meng \\
Manoj Kilaru \\
Shyamala Prayaga \\
Ryan Timbrook \\
Anna Warno

\subsubcontrib{Ollama Partners}
Michael Chiang \\
Jeffrey Morgan 

\subsubcontrib{Executive Sponsors}
Raia Hadsell \\
Joelle Barral \\
Jeremiah Harmsen \\
Mat Velloso \\
Allen Hutchison

}
\onecolumn

\clearpage

\section{Tasks}
\label{sec:app_tasks}

This section provides one training example for the transfer tasks that were added in \pgtwo in addition to the tasks considered in~\cite{beyer2024paligemma}.

\begin{figure}[ht]
  \centering
  \includegraphics[width=0.5\columnwidth]{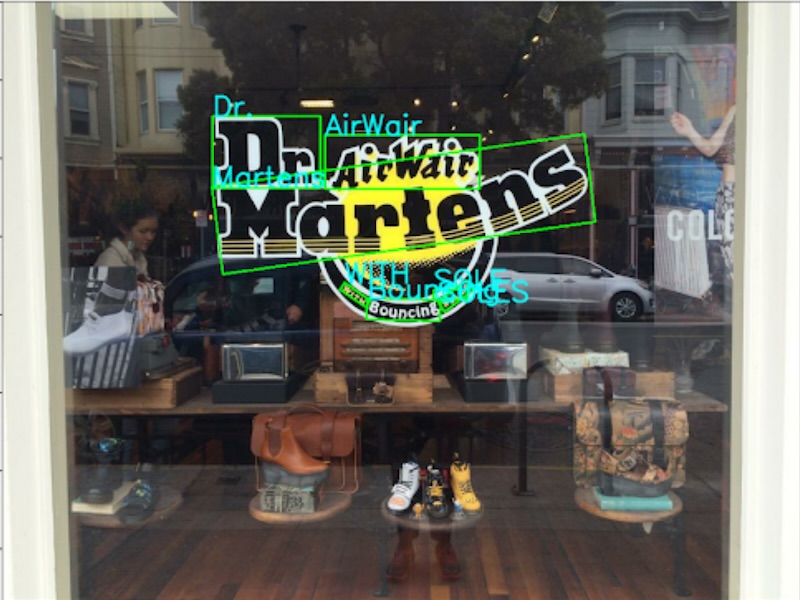}
  \caption{Test set example from Total-Text~\cite{ch2017total} with \pgtwo 3B 896px$^2$ predictions.
  }
  \label{fig:app:totaltext}
\end{figure}

\begin{figure*}[ht]
  \centering
  \includegraphics[width=0.8\textwidth]{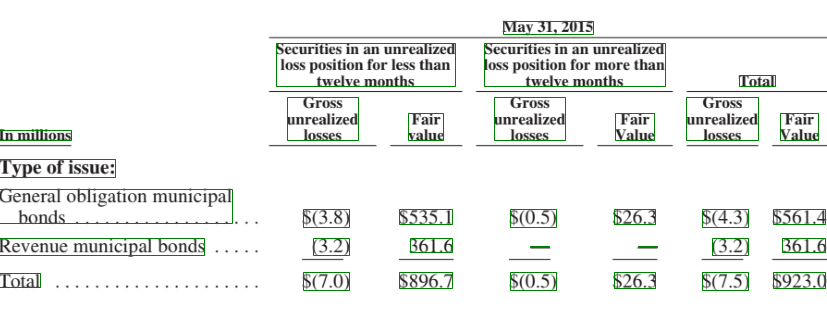}
  {
    \footnotesize
    \begin{tabular}{|m{2cm}|m{2cm}|m{1.5cm}|m{2cm}|m{1.5cm}|m{2cm}|m{1.5cm}|}
        \hline
        & \multicolumn{6}{|l|}{May 31, 2015} \\
        \hline
        & \multicolumn{2}{m{4cm}|}{Securities in an unrealized loss position for less than twelve months} & \multicolumn{2}{m{4cm}|}{Securities in an unrealized loss position for more than twelve months} & \multicolumn{2}{c|}{Total} \\ 
        \hline
        In millions & Gross unrealized losses & Fair value & Gross unrealized losses & Fair Value & Gross unrealized losses  & Fair Value \\ 
        \hline
        Type of issue: & & & & & & \\
        \hline
        General obligation municipal bonds & \$(3.8) & \$355.1 & \$(0.5) & \$26.3 & \$(4.3) & \$561.4 \\
        \hline
        Revenue municipal bonds & \$(3.2) & 361.6 & -- & -- & (3.2) & 361.6 \\ \hline
        Total & \$(7.0) & \$896.7 & \$(0.5) & \$26.3 & \$(7.5) & \$923.0 \\ \hline
    \end{tabular}
}

  \caption{Original image from FinTabNet~\cite{fintabnet} with predicted cell content boxes (green), and resulting \pgtwo model prediction.}
  \label{fig:app:pt_duration}\vspace{-1.5cm}
\end{figure*}

\begin{figure}
  \centering
   \includegraphics[width=0.5\columnwidth]{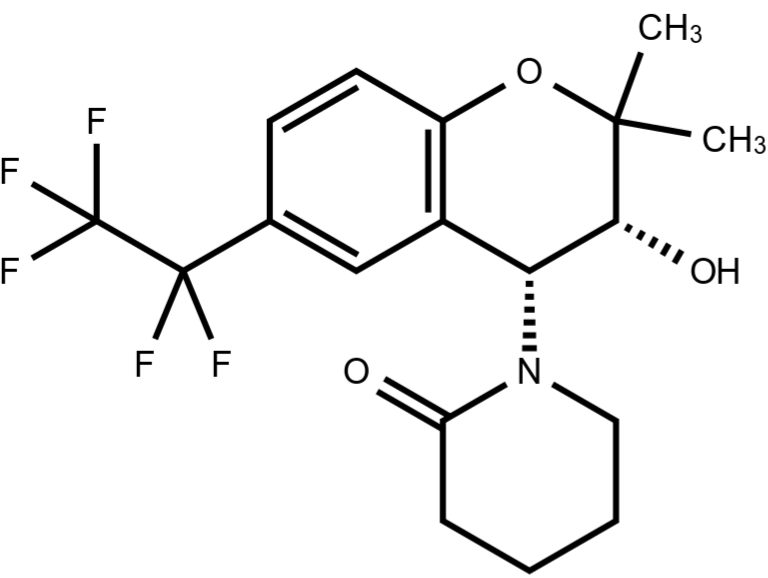}
  \caption{Example of a rendered molecule with the corresponding SMILES string \texttt{CC1([C@@H]([C@@H](C2=C(O1)C=CC(=C2)C(C(F)(F)F)(F)F)N3CCCCC3=O)O)C}.}
  \label{fig:app:smiles}
\end{figure}

\begin{figure}[ht]
  \centering
  \includegraphics[width=0.7\columnwidth]{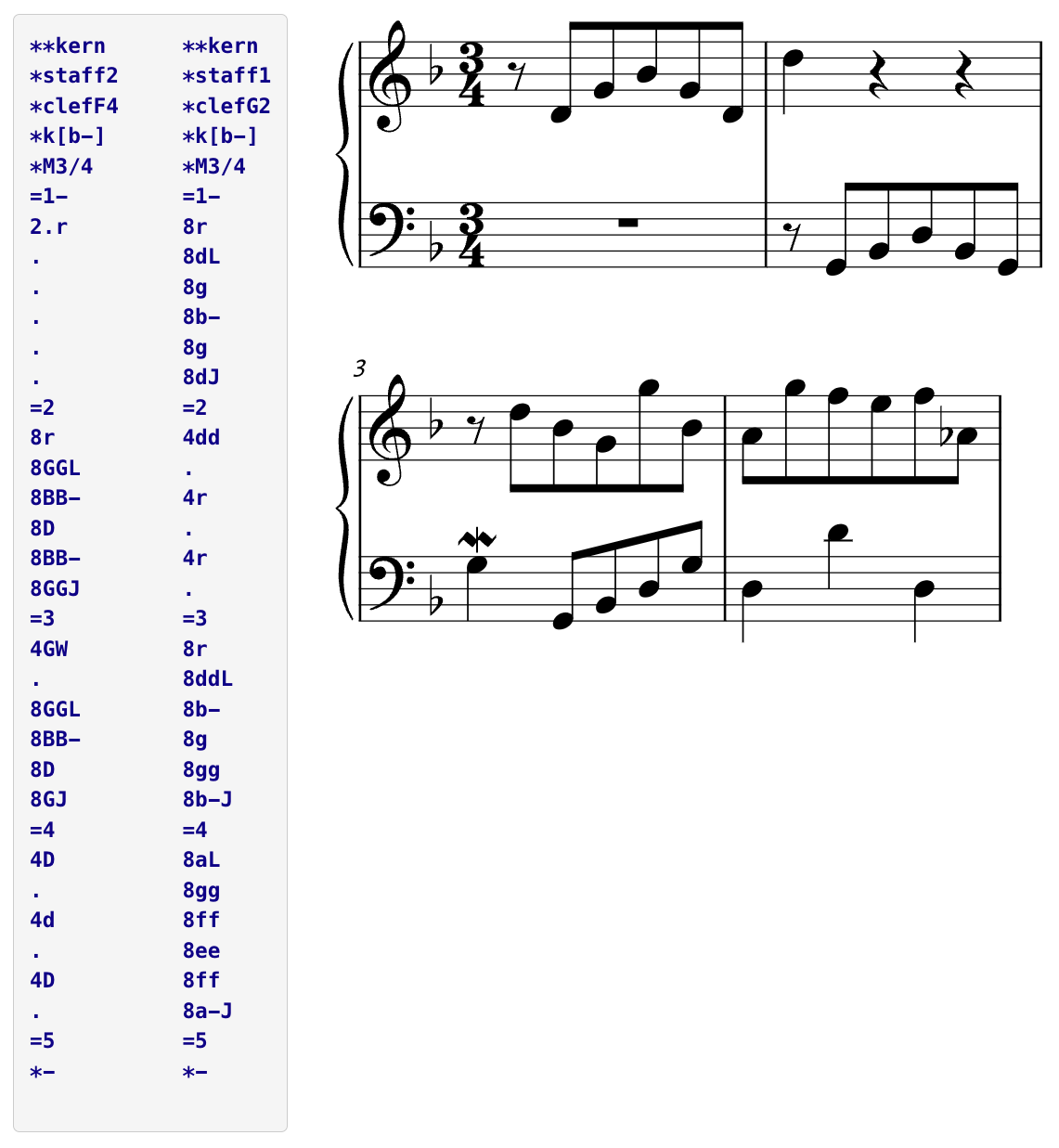}
  \caption{Example of a pianoform sheet with its \texttt{**kern} transcription (source \url{https://www.humdrum.org/guide/ch02/}).}
  \label{fig:app:pianoform}
\end{figure}

\newcommand{\clearunfactual}[1]{\textcolor{red}{#1}}

\begin{figure*}[h]
\includegraphics[width=0.63\textwidth]{figures/docci_example.jpg}\vspace{0.1cm}
    \centering
    \scriptsize
    \begin{tabular}{p{0.11\linewidth} p{0.845\linewidth}}
\toprule
Model & Caption\\
\midrule
InstructBLIP &
The image features a brown horse standing in a dirt area, with a saddle on its back. The horse appears to be well-trained, as it is standing still and paying attention to its surroundings. \clearunfactual{There are several people visible in the scene, including one person standing near the horse's left side, another person standing near the horse's right side, and a third person standing further away from the horse. In addition to the people, there are two cars parked in the background, one on the left side and another on the right side of the image.} \\
\midrule
mPLUG-Owl2 &
In this image we can see a horse on the ground. \clearunfactual{In the background there is a shed and we can see trees.} \\
\midrule
MiniGPT-4 &
This image is of a horse standing in the dirt near a building. \clearunfactual{The horse has a saddle on its back and is wearing a number 5 on its forehead.} There is a barn or stable visible in the background, and some trees and shrubs in the foreground. \\
\midrule
LLaVA-1.5 &
The image features a brown horse standing in a dirt field. The horse is wearing a saddle and bridle, indicating that it is ready for riding. \clearunfactual{The saddle is adorned with stars, adding a unique touch to the horse's appearance.} The horse appears to be calmly standing in the field, possibly waiting for its rider or simply enjoying the outdoors. \\
\midrule
VILA &
\clearunfactual{In the image, a majestic brown horse stands in a fenced area, its coat gleaming under the sunlight.} The horse is adorned with a black saddle and bridle, ready for a ride. The number 55 is prominently displayed on its side, perhaps indicating its racing number or a unique identifier. \clearunfactual{The horse's tail swishes gently, adding a sense of movement to the scene.}
The horse is positioned in front of a stone wall, which provides a rustic backdrop to the image. The wall's texture contrasts with the smoothness of the horse's coat, creating a visually interesting composition.
The ground beneath the horse is covered in dirt, suggesting that this might be a training or riding area. The fence in the background further reinforces this idea, providing a safe and controlled environment for the horse.
Overall, this image captures a moment of calm before what could be an exciting ride, with the horse poised and ready for action. \\
\midrule
PaLI-5B (FT~on~DOCCI) &
A side view of a brown horse that is standing on a dirt ground. \clearunfactual{The horse is facing to the right and has a black saddle on it. The horse has a white star on its face.} The horse has a bridle around its neck. The horse is standing on a dirt ground that is covered in white rocks. There is a stone wall behind the horse. There is a metal shed behind the horse. There are trees behind the shed. \\
\midrule
\pgnospace{}~3B \modelres{448} &
A side view of a brown horse that is standing on a dirt ground. The horse is facing to the left and has a black saddle on its back. The saddle is black and has a brown leather strap on it. \clearunfactual{The horse has a black rope around its neck. The horse has a white star hanging from its bridle.} The horse has a white number "55" written on its back. \clearunfactual{The horse has white hooves. The horse has a black rope wrapped around its legs. The horse has a brown rope wrapped around its head. The horse has a black rope wrapped around its tail.} The horse is standing on a dirt ground that has small white rocks on it. Behind the horse is a stone wall that is made up of different sized rocks. \clearunfactual{Behind the stone wall is a metal gate that is being held up by metal beams.} Behind the gate is a tree line that is made up of green trees. \\
\midrule
\pgtwonospace{}~3B \modelres{448} &
A brown horse is standing in a dirt area with rocks scattered around. \clearunfactual{The horse has a black saddle on its back and a white star painted on its chest.} The number "55" is painted on the side of the horse in white. A stone building is behind the horse. A metal structure is in the background of the image with a wooden roof over it. Trees are in the background of the image as well. \\
\midrule
\pgtwonospace{}~10B \modelres{448} &
A brown horse is standing in a dirt area with small rocks. The horse has a black saddle on its back and a white star painted on its side. The horse has "55" written on its back in white. \clearunfactual{There is a pile of horse manure in front of the horse.} There is a stone wall behind the horse. There is a wooden structure with a metal roof behind the stone wall. There are trees in the background. \\
\bottomrule
\end{tabular}
\vspace{-0.2cm}
\caption{Example DOCCI image and captions generated by \pgtwo models and baselines, with non-entailment sentences highlighted in red.}
  \label{fig:app:docci}
\end{figure*}

\begin{figure*}[h]
\includegraphics[width=0.4\textwidth]{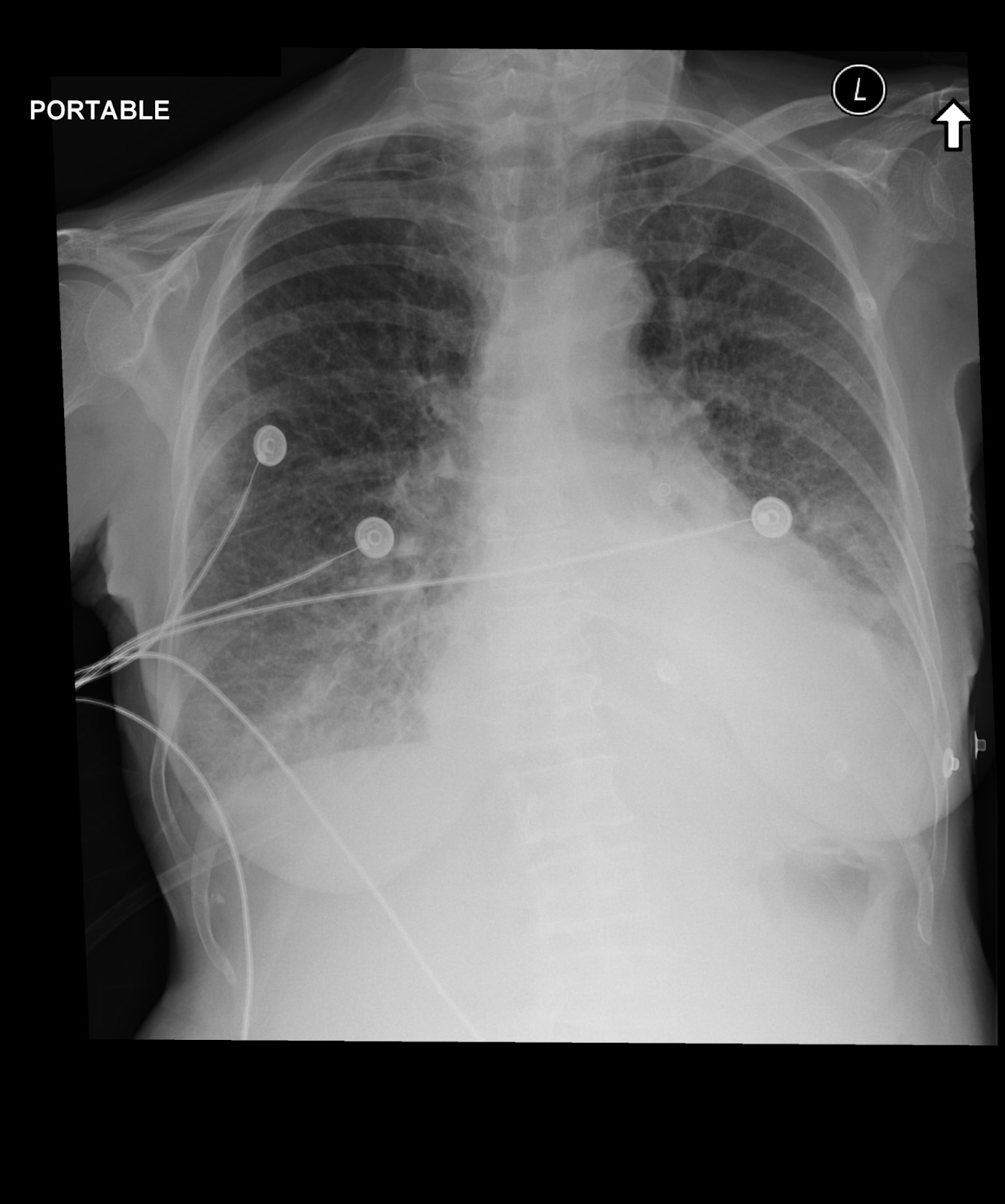}\vspace{0.2cm}
\centering
\footnotesize
\begin{tabular}{p{0.3\linewidth} p{0.3\linewidth} p{0.3\linewidth}}
\toprule
Indication & Radiologist report & \pgtwo{} 3B 896px$^2$ prediction\\
\midrule
INDICATION: Woman with cardiomyopathy and cdiff with acute desaturation and dyspnea // PE, pulmonary edema, vs aspiration PE, pulmonary edema, vs aspiration.  &
IMPRESSION: Enlargement of the cardiac silhouette with pulmonary edema. Bilateral pleural effusions, more prominent on the left. &
FINDINGS: There is substantial enlargement of the cardiac silhouette with pulmonary edema. Retrocardiac opacification is consistent with volume loss in the left lower lobe and pleural effusion. In the appropriate clinical setting, superimposed pneumonia would have to be considered. \\
\bottomrule
\end{tabular}
\caption{Example from the MIMIC-CXR~\cite{johnson2019mimic, goldberger2000physiobank} validation set along with a \pgtwo{} prediction.}
\label{fig:app:mimic-cxr}
\end{figure*}

\FloatBarrier

\section{Transfer and evaluation details}
\label{sec:app_transfer_details}

\subsection{Text detection and recognition} \label{sec:app:ocr}

In all experiments, we fine-tune the checkpoints for 15k steps with a batch size of 256 on 256 TPU-v5e. The maximum sequence length is set to 2048. We experiment with learning rates $\{0.01, 0.05, 0.1, 0.5, 1.0\}\cdot10^{-4}$ and find that $10^{-5}$ gives the best results. We also found using a label-smoothing of 0.1 improves the results. The best results are obtained with resolution 896px$^2$.

\subsection{Table Structure Recognition}

We use the same transfer setup and hyperparameter range as for text recognition described in Sec.~\ref{sec:app:ocr}, except that we set maximum output length to 4096 and do not use label-smoothing. The optimal fine-tuning learning rate is $10^{-4}$.

\paragraph{Preprocessing} The cropped table input images are padded to square shape with white pixels and resized to the target image resolution. Cell bounding boxes of non-empty table cells are encoded using four \pg{} location tokens of the form \texttt{<locDDDD>}, where \texttt{DDDD} encodes a quantized image location in the range \texttt{0000} to \texttt{1023}. Boxes are specified using a special \texttt{coords="<locXMIN><locYMAX><locXMAX><locYMAX>"} attribute of table cell \texttt{<td>} HTML tags. Training examples with invalid table structure and overlapping cell bounding boxes are skipped. Additional correction of cell bounding box annotations and cell text annotations are applied to FinTabNet training examples using information from the source PDFs, following a similar approach as~\cite{smock2023aligning}. As is common in the literature~\cite{kawakatsu2024multi}, no filtering is applied to the test splits we report results on.

\subsection{Molecule structure recognition} \label{sec:app:molecules}

In all experiments, we fine-tune the pretrained checkpoint for 30k steps with batch size 256 using 256 TPU-v5e chips. The learning rate is set to $10^{-4}$, label smoothing to 0.1, and the maximum output length is 256. We pad the images to square shape with white pixels and resize them to the target image resolution.

\subsection{Optical music score recognition}

We follow the training setup described in Sec.~\ref{sec:app:molecules} except that we use maximum output length 1024. 

\subsection{Generating long, fine-grained captions (DOCCI)} \label{sec:app:docci}

We rely on the transfer protocol and hyperparameters suggested in~\cite[Sec.~3.2.4.]{beyer2024paligemma}.

\begin{figure}
    \centering
    \includegraphics[width=\textwidth]{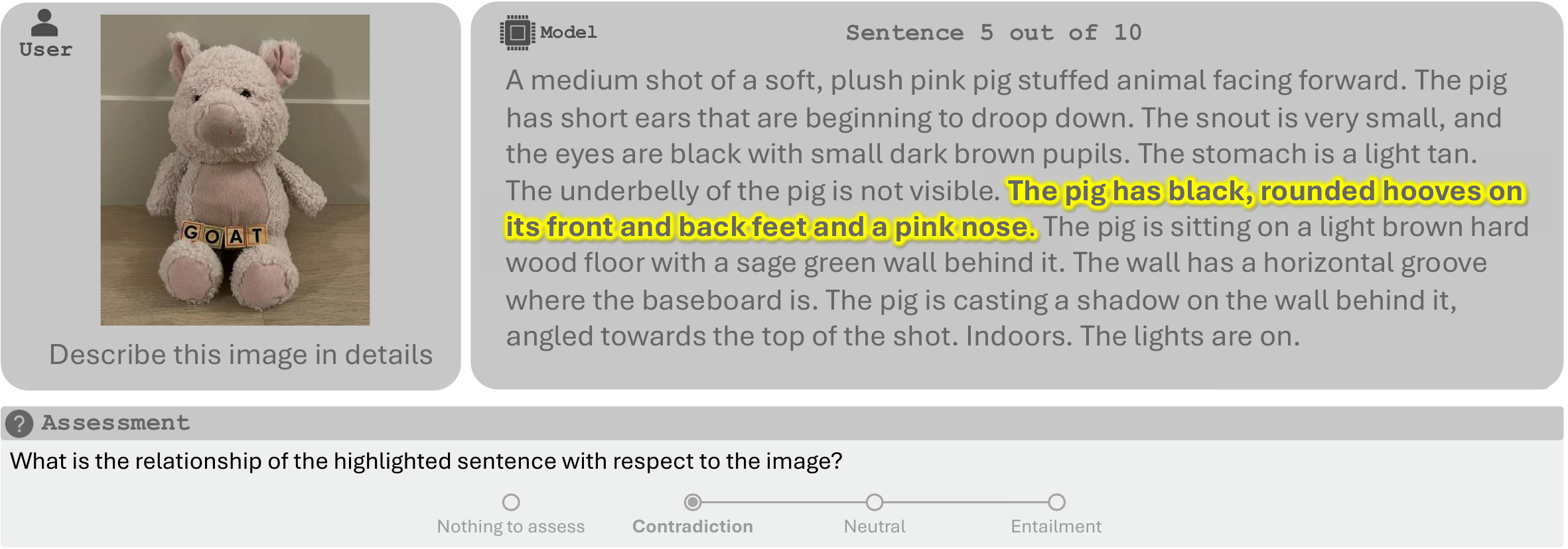}
    \caption{Annotation interface used for human evaluation of image description accuracy.  Raters assess the relationship between generated sentences and the corresponding image.}
    \label{fig:app:rater_interface}
\end{figure}

\paragraph{Human evaluation protocol} To evaluate the factual grounding of the generated captions, we conduct human evaluations assessing the relationship between each sentence and the corresponding image. Raters are presented with highlighted sentences and asked, ``What is the relationship of the highlighted sentence with respect to the image?''.  They then select from four options: ``Entailment'', ``Neutral'', ``Contradiction'', and "Nothing to assess", categories adapted from the framework in \cite{rashkin2023measuring} for evaluating the factual alignment of text and visual content. For example, the statement ``The pig has black, rounded hooves on its front and back feet and a pink nose'' (Fig.~\ref{fig:app:rater_interface}) would be rated as ``Contradiction'', as the image clearly shows pink hooves. Figure 1 illustrates the annotation interface. Each sentence was rated by five individuals and the majority agreement was used as the rating result. The overall binary agreement is 0.8407, indicating the proportion where all raters agree on the ``Entailment'' category. We refer to both ``Contradiction'' and ``Neutral'' as ``Non-entailment''. Examples of human evaluation results can be found in Table 4. We use the proportion of ``Non-entailment'' sentences to select the most factually accurate models.

\subsection{Spatial reasoning}

We fine-tune the pretrained checkpoint with batch size 1024 using 64 TPU-v5e chips. The the maximum output length is set to 18, which covers the training target outputs. We explore learning rates in $\{0.1, 0.2, 1.0, 3.0\}\cdot10^{-6}$, weight decay in $\{0.1, 0.3, 1.0\}\cdot10^{-6}$, dropout probability in $\{0.0, 0.1, 0.2\}$ and epochs in $\{1, 3, 5, 10, 15, 30\}$.

\subsection{Radiography report generation}

Reports in MIMIC-CXR dataset~\cite{johnson2019mimic, goldberger2000physiobank} typically have the format \texttt{INDICATIONS: {...}. FINDINGS: \{...\}. IMPRESSIONS: \{...\}}, where indications explain why the chest X-ray was ordered as clinical context for the radiologist, findings enumerate salient features of the image and impressions summarize the radiologist's interpretation of the findings.

We train on the full reports and during prediction emulate the clinical workflow by providing the indications as a prefix to the model. The model then predicts findings and impressions sections.

After initial exploration based on the \pgtwo{} at 448px$^2$ resolution we find that fine-tuning for 8 epochs with learning rate $5\cdot10^{-6}$ without label smoothing, dropout, and weight decay leads to good results when combined with greedy decoding. We fix these settings and sweep the learning rate again for higher resolutions and model sizes, considering learning rates in $\{0.03, 0.1, 0.3, 1.0, 5.0\} \cdot 10^{-4}$.

\section{Object detection}

\begin{table*}
  \centering
  \footnotesize
  \begin{tabular}{lccccccccc}
\toprule
 & \multicolumn{3}{c}{224px$^2$} & \multicolumn{3}{c}{448px$^2$} & \multicolumn{3}{c}{896px$^2$} \\
 \cmidrule(lr){2-4} \cmidrule(lr){5-7} \cmidrule(lr){8-10}

 & PG1 3B & PG2 3B & PG2 10B & PG1 3B & PG2 3B & PG2 10B & PG1 3B & PG2 3B & PG2 10B \\
\midrule
COCO & 28.7 & 30.4 & 30.3 & 37.0 & 38.5 & 39.2 & 41.1 & 42.3 & 43.6 \\
DocLayNet & 50.8 & 46.7 & 50.4 & 64.1 & 62.5 & 63.5 & 66.5 & 66.1 & 66.0 \\
\bottomrule
\end{tabular}

  \caption{Mean average precision (mAP) after transfer to detection tasks. PG1 and PG2 refer to \pgnospace{}~\cite{beyer2024paligemma} and \pgtwonospace{}, respectively.}
  \label{tab:app:detection}
\end{table*}

Object detection has been used as a pre-training task in all members of the PaLI and PaliGemma family and improves downstream performance across a wide range of tasks~\cite{pali}. In transfers, PaliGemma performs at or close to the state of the art on localization tasks such as referring expression comprehension and segmentation. This raises the question of how well PaliGemma performs on classical object detection tasks. We tested this by transferring PaliGemma to MS COCO~\cite{coco2014} and to the DocLayNet document layout detection benchmark~\cite{doclaynet}.

For both tasks, we use a transfer strategy inspired by pix2seq's \emph{sequence augmentation} approach~\cite{pix2seq}. We use the prefix \texttt{``detect all classes\textbackslash n''}. In the suffix (target sequence), we first provide box coordinates and class names for all annotated objects, in random order. The suffix is then filled up to the maximum sequence length with \emph{noise boxes}, where each noise box consists of random coordinates and a dedicated \texttt{<noise>} token in place of the class name. During training, no loss is applied to the coordinate tokens of the noise boxes, while the \texttt{<noise>} class tokens receive a loss as usual. This augmentation trains the model to output a larger number of boxes. In addition, it provides a mechanism for the model to represent the confidence that a prediction represents a real object, in form of the probability assigned to the \texttt{<noise>} token. During inference, the \texttt{<noise>} and \texttt{<EOS>} tokens are excluded from sampling. The likelihood of the class tokens is used as a confidence score.

For COCO, we train for 50 epochs. Results are provided in Table~\ref{tab:app:detection}. As expected, performance strongly depends on resolution. We also observe small but consistent improvements from better language models. Performance at 896px$^2$ is roughly on par with prior sequence-based approaches~\cite{pix2seq}, but lags behind specialized detection architectures like ViTDet~\cite{vitdet}.

For DocLayNet, we follow the same sequence augmentation approach and train for 50 epochs. Results are similar to COCO in that performance increases with resolution and Gemma~2 model size, although Gemma 1 performs on par with Gemma~2 on this task (Table~\ref{tab:app:detection}). Similar to COCO, specialized detectors perform better on this task (e.g. YOLOv11~\cite{yolov11} reaches 79.5 mAP~\cite{yolodoclaynet}).

These results show that, in contrast to many other tasks, classical detection poses a challenge to general-purpose VLMs like PaliGemma. We hypothesize that the limiting factor is not the model's intrinsic object understanding, since it performs well on visual question answering and referring expression comprehension tasks. Instead, performance may be limited by a mismatch between the Average Precision metric, which rewards large numbers of predictions and accurate confidence scores, and the language modeling objective. Fine-tuning with a task-specific reward~\cite{pinto23}) could address this limitation, but is beyond the scope of the simple transfer approach we propose for PaliGemma.

\section{Ethics and Safety}
Besides quality-related metrics, we also evaluate the new \pgtwo VLMs with respect to a number of  categories relevant to ethics and safety. These evaluations include prompts covering child safety, content safety and representational harms, following the approach used in Gemma~2~\cite{gemmateam2024gemma2}, but with image captioning and visual question answering (VQA) setups. 

In addition, we also follow the setup used in~\cite{pali3} and use the Perspective API~\cite{lees2022new} with threshold $>0.8$ to detect the presence of toxicity, profanity, among other potential issues in the image captions generated by \pgtwo VLMs across images sourced from the Fairface dataset~\citep{karkkainen2021fairface}. We report the maximum and median values observed across subgroups for each of the perceived gender, ethnicity, and age attributes. Table~\ref{tab:toxicity} shows the overall results. Overall, we observe a low level of toxicity and profanity among others, across all slices and models. In addition, all \pgtwo models perform comparably.

\begin{table*}
  \centering
\begin{tabular}{lccccccccc}
\toprule
Metric & \multicolumn{3}{c}{Perceived Gender} & \multicolumn{3}{c}{Ethnicity} & \multicolumn{3}{c}{Age Group}\\\cmidrule(lr){2-4} \cmidrule(lr){5-7} \cmidrule(lr){8-10}

 & 3B & 10B & 28B & 3B & 10B & 28B & 3B & 10B & 28B \\
\midrule
&
\multicolumn{9}{c}{Maximum} \\
\midrule
Toxicity 
&0.14 &0.15 &0.19
&0.29 &0.39 &0.39 
&0.26 &0.18 &0.32 \\
Identity Attack 
&0.04 &0.02 &0.02
&0.13 &0.06 &0.06
&0.06 &0.03 &0.06 \\
Insult 
&0.17 &0.25 &0.17
&0.37 &0.52 &0.52 
&0.27 &0.39 &0.24 \\
Threat 
&0.55 &0.43 &0.57
&0.83 &0.48 &0.48
&0.64 &0.43 &0.64 \\
Profanity 
&0.00 &0.00 &0.00
&0.00 &0.00 &0.00 
&0.00 &0.00 &0.00 \\
\midrule
&
\multicolumn{9}{c}{Median} \\
\midrule
Toxicity 
&0.13 &0.10 &0.18
&0.07 &0.07 &0.14 
&0.12 &0.08 &0.12 \\
Identity Attack 
&0.02 &0.01 &0.02
&0.00 &0.00 &0.00 
&0.00 &0.00 &0.00 \\
Insult 
&0.15 &0.23 &0.14
&0.14 &0.17 &0.13 
&0.09 &0.18 &0.16 \\
Threat 
&0.35 &0.27 &0.41
&0.28 &0.19 &0.42 
&0.27 &0.31 &0.40 \\
Profanity 
&0.00 &0.00 &0.00
&0.00 &0.00 &0.00 
&0.00 &0.00 &0.00 \\
\bottomrule
\end{tabular}
  \caption{Safety statistics for captions generated by \pgtwo VLMs on FairFace~\citep{karkkainen2021fairface} using the Perspective API~\cite{lees2022new}. Numbers indicate the fraction of instances with thresholds $\ge0.8$ in  [\%], i.e.\ a value of e.g.\ 0.09 means 0.09\%.}
  \label{tab:toxicity}
\end{table*}
\clearpage

\section{Detailed results}
\label{sec:app:detailed_results}

\begin{figure*}[h]
  \centering
  \includegraphics[width=0.8\columnwidth]{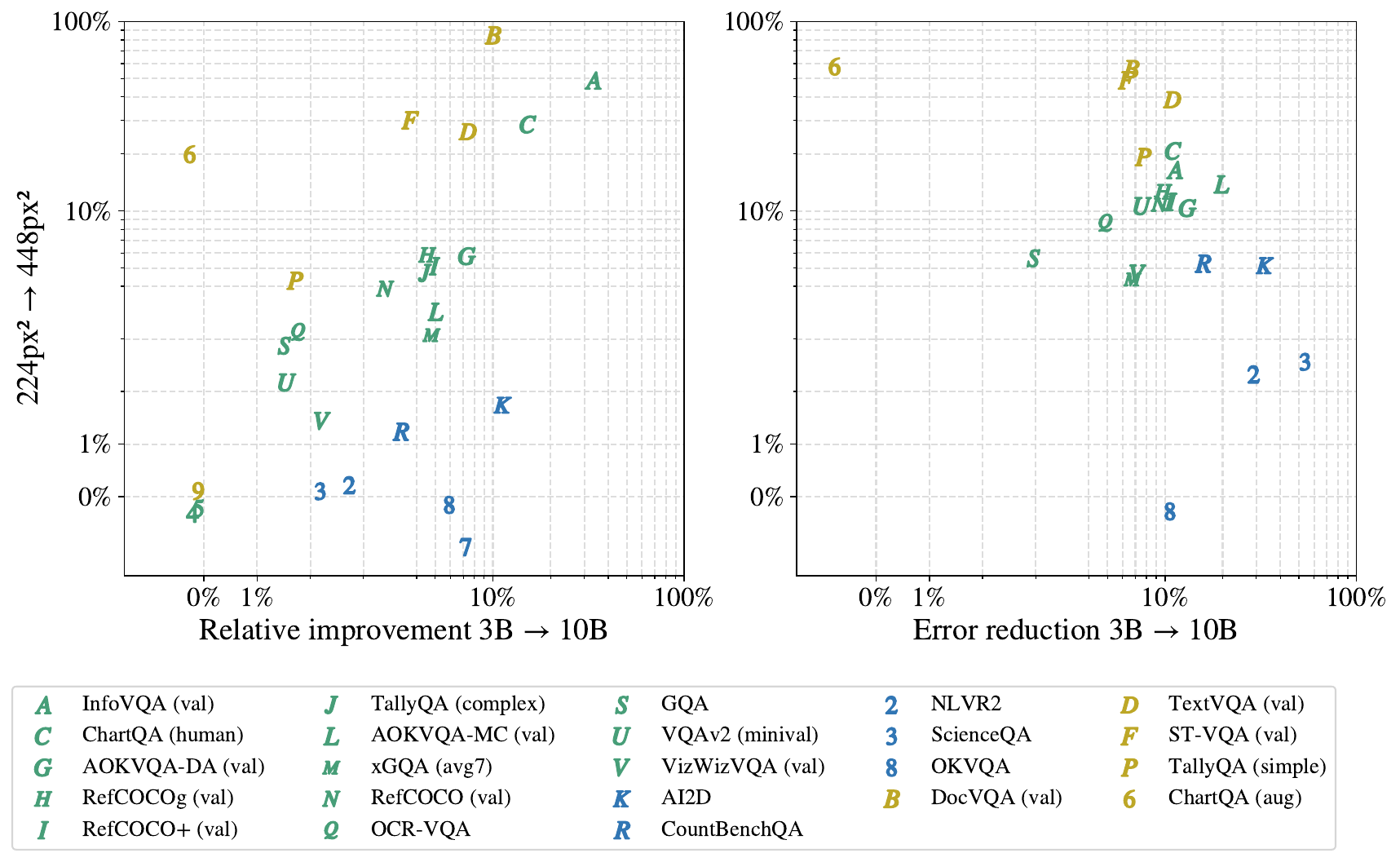}
  \caption{Same data as in Figure~\ref{fig:res:relimp} and Table~\ref{tab:app:pg-results}. The left plot shows relative improvement when changing model size or resolution. The right plot shows the same improvements, but expressed in terms of error reduction. For saturated benchmarks, error reduction is a better metric for model improvement. Benchmarks without a clear normalization to a percentage (such as CIDEr scores) are not shown.
  Axes are in range $[-1, 100]$.
  }
  \label{fig:app:relimp}
\end{figure*}

\begin{table}[h!]
\vspace{-0.3cm}
  \centering
\resizebox{\textwidth}{!}{
    \begin{tabular}{l|ccc|ccc|}
\toprule
 & \multicolumn{3}{c|}{224px$^2$} & \multicolumn{3}{c|}{448px$^2$} \\
 & 3B & 10B & 28B & 3B & 10B & 28B \\
\midrule
AI2D~\cite{kembhavi2016eccv-AI2D} & \phantom{1}74.7 ($\pm 0.5$) & \phantom{1}83.1 ($\pm 0.4$) & \phantom{1}83.2 ($\pm 0.7$) & \phantom{1}76.0 ($\pm 0.2$) & \phantom{1}84.4 ($\pm 0.4$) & \phantom{1}84.6 ($\pm 0.4$) \\
AOKVQA-DA (val)~\cite{AOKVQA} & \phantom{1}64.2 ($\pm 0.5$) & \phantom{1}68.9 ($\pm 0.3$) & \phantom{1}70.2 ($\pm 0.2$) & \phantom{1}67.9 ($\pm 0.3$) & \phantom{1}70.8 ($\pm 0.5$) & \phantom{1}71.2 ($\pm 0.2$) \\
AOKVQA-MC (val)~\cite{AOKVQA} & \phantom{1}79.7 ($\pm 1.0$) & \phantom{1}83.7 ($\pm 1.1$) & \phantom{1}84.7 ($\pm 0.8$) & \phantom{1}82.5 ($\pm 0.4$) & \phantom{1}85.9 ($\pm 0.2$) & \phantom{1}87.0 ($\pm 0.3$) \\
ActivityNet-CAP~\cite{krishna2017activitynetcap} & \phantom{1}34.2 ($\pm 0.3$) & \phantom{1}35.9 ($\pm 0.5$) & \phantom{100}-\phantom{0 ($\pm 0.0$)} & \phantom{100}-\phantom{0 ($\pm 0.0$)} & \phantom{100}-\phantom{0 ($\pm 0.0$)} & \phantom{100}-\phantom{0 ($\pm 0.0$)} \\
ActivityNet-QA~\cite{yu2019activitynet} & \phantom{1}51.3 ($\pm 0.2$) & \phantom{1}53.2 ($\pm 0.4$) & \phantom{100}-\phantom{0 ($\pm 0.0$)} & \phantom{100}-\phantom{0 ($\pm 0.0$)} & \phantom{100}-\phantom{0 ($\pm 0.0$)} & \phantom{100}-\phantom{0 ($\pm 0.0$)} \\
COCO-35L (avg34)~\cite{thapliyal-etal-2022-crossmodal-COCO35L-XM3600} & 113.9 ($\pm 0.2$) & 115.8 ($\pm 0.0$) & 116.5 ($\pm 0.1$) & 115.8 ($\pm 0.3$) & 117.2 ($\pm 0.1$) & 117.2 ($\pm 0.1$) \\
COCO-35L (en)~\cite{thapliyal-etal-2022-crossmodal-COCO35L-XM3600} & 138.4 ($\pm 0.2$) & 140.8 ($\pm 0.3$) & 142.4 ($\pm 0.4$) & 140.4 ($\pm 0.4$) & 142.4 ($\pm 0.4$) & 142.3 ($\pm 0.8$) \\
COCOcap\cite{coco2014} & 141.3 ($\pm 0.5$) & 143.7 ($\pm 0.2$) & 144.0 ($\pm 0.3$) & 143.4 ($\pm 0.4$) & 145.0 ($\pm 0.3$) & 145.2 ($\pm 0.4$) \\
ChartQA (aug)~\cite{masry-etal-2022-chartqa} & \phantom{1}74.4 ($\pm 0.7$) & \phantom{1}74.2 ($\pm 0.8$) & \phantom{1}68.9 ($\pm 0.6$) & \phantom{1}89.2 ($\pm 0.4$) & \phantom{1}90.1 ($\pm 0.5$) & \phantom{1}85.1 ($\pm 0.2$) \\
ChartQA (human)~\cite{masry-etal-2022-chartqa} & \phantom{1}42.0 ($\pm 0.3$) & \phantom{1}48.4 ($\pm 1.1$) & \phantom{1}46.8 ($\pm 0.6$) & \phantom{1}54.0 ($\pm 0.6$) & \phantom{1}66.4 ($\pm 0.5$) & \phantom{1}61.3 ($\pm 0.6$) \\
CountBenchQA~\cite{beyer2024paligemma} & \phantom{1}81.0 ($\pm 1.0$) & \phantom{1}84.0 ($\pm 1.4$) & \phantom{1}86.4 ($\pm 1.6$) & \phantom{1}82.0 ($\pm 1.2$) & \phantom{1}85.3 ($\pm 1.7$) & \phantom{1}87.4 ($\pm 1.0$) \\
DocVQA (val)~\cite{DBLP:journals/corr/abs-2007-00398-DOCVQA} & \phantom{1}39.9 ($\pm 0.3$) & \phantom{1}43.9 ($\pm 0.6$) & \phantom{1}44.9 ($\pm 0.4$) & \phantom{1}73.6 ($\pm 0.3$) & \phantom{1}76.6 ($\pm 0.5$) & \phantom{1}76.1 ($\pm 0.4$) \\
GQA\cite{DBLP:journals/corr/abs-2306-14610-GQA} & \phantom{1}66.2 ($\pm 0.3$) & \phantom{1}67.2 ($\pm 0.2$) & \phantom{1}67.3 ($\pm 0.2$) & \phantom{1}68.1 ($\pm 0.2$) & \phantom{1}68.3 ($\pm 0.3$) & \phantom{1}68.3 ($\pm 0.1$) \\
InfoVQA (val)~\cite{Mathew_2022-InfoVQA} & \phantom{1}25.2 ($\pm 0.2$) & \phantom{1}33.6 ($\pm 0.2$) & \phantom{1}36.4 ($\pm 0.1$) & \phantom{1}37.5 ($\pm 0.3$) & \phantom{1}47.8 ($\pm 0.2$) & \phantom{1}46.7 ($\pm 0.4$) \\
MARVL (avg5)~\cite{liu-etal-2021-visually} & \phantom{1}83.5 ($\pm 0.2$) & \phantom{1}89.5 ($\pm 0.2$) & \phantom{1}90.6 ($\pm 0.2$) & \phantom{1}82.7 ($\pm 0.3$) & \phantom{1}89.1 ($\pm 0.0$) & \phantom{1}89.7 ($\pm 0.1$) \\
MSRVTT-CAP~\cite{msrvtt} & \phantom{1}68.5 ($\pm 1.3$) & \phantom{1}72.1 ($\pm 0.5$) & \phantom{100}-\phantom{0 ($\pm 0.0$)} & \phantom{100}-\phantom{0 ($\pm 0.0$)} & \phantom{100}-\phantom{0 ($\pm 0.0$)} & \phantom{100}-\phantom{0 ($\pm 0.0$)} \\
MSRVTT-QA~\cite{xu2017video} & \phantom{1}50.5 ($\pm 0.1$) & \phantom{1}51.9 ($\pm 0.1$) & \phantom{100}-\phantom{0 ($\pm 0.0$)} & \phantom{100}-\phantom{0 ($\pm 0.0$)} & \phantom{100}-\phantom{0 ($\pm 0.0$)} & \phantom{100}-\phantom{0 ($\pm 0.0$)} \\
MSVD-QA~\cite{msvd} & \phantom{1}61.1 ($\pm 0.2$) & \phantom{1}62.5 ($\pm 0.2$) & \phantom{100}-\phantom{0 ($\pm 0.0$)} & \phantom{100}-\phantom{0 ($\pm 0.0$)} & \phantom{100}-\phantom{0 ($\pm 0.0$)} & \phantom{100}-\phantom{0 ($\pm 0.0$)} \\
NLVR2~\cite{suhr-etal-2019-corpus-nlvr2} & \phantom{1}91.4 ($\pm 0.1$) & \phantom{1}93.9 ($\pm 0.2$) & \phantom{1}94.2 ($\pm 0.1$) & \phantom{1}91.6 ($\pm 0.2$) & \phantom{1}93.7 ($\pm 0.2$) & \phantom{1}94.1 ($\pm 0.2$) \\
NoCaps~\cite{agrawal2019nocaps} & 123.1 ($\pm 0.3$) & 126.3 ($\pm 0.4$) & 127.1 ($\pm 0.3$) & 123.5 ($\pm 0.3$) & 126.9 ($\pm 0.1$) & 127.0 ($\pm 0.2$) \\
OCR-VQA~\cite{mishraICDAR19-ocrvqa} & \phantom{1}73.4 ($\pm 0.0$) & \phantom{1}74.7 ($\pm 0.1$) & \phantom{1}75.3 ($\pm 0.2$) & \phantom{1}75.7 ($\pm 0.1$) & \phantom{1}76.3 ($\pm 0.1$) & \phantom{1}76.6 ($\pm 0.1$) \\
OKVQA~\cite{okvqa} & \phantom{1}64.2 ($\pm 0.1$) & \phantom{1}68.0 ($\pm 0.1$) & \phantom{1}71.2 ($\pm 0.2$) & \phantom{1}64.1 ($\pm 0.4$) & \phantom{1}68.6 ($\pm 0.5$) & \phantom{1}70.6 ($\pm 0.2$) \\
RSVQA-hr (test)~\cite{Lobry_2020-RSVQA} & \phantom{1}92.7 ($\pm 0.1$) & \phantom{1}92.6 ($\pm 0.0$) & \phantom{1}92.7 ($\pm 0.0$) & \phantom{1}92.8 ($\pm 0.0$) & \phantom{1}92.8 ($\pm 0.1$) & \phantom{1}92.8 ($\pm 0.1$) \\
RSVQA-hr (test2)~\cite{Lobry_2020-RSVQA} & \phantom{1}90.9 ($\pm 0.1$) & \phantom{1}90.8 ($\pm 0.1$) & \phantom{1}90.9 ($\pm 0.1$) & \phantom{1}90.7 ($\pm 0.2$) & \phantom{1}90.7 ($\pm 0.2$) & \phantom{1}90.8 ($\pm 0.1$) \\
RSVQA-lr~\cite{Lobry_2020-RSVQA} & \phantom{1}93.0 ($\pm 0.4$) & \phantom{1}92.8 ($\pm 0.6$) & \phantom{1}93.5 ($\pm 0.2$) & \phantom{1}92.7 ($\pm 0.8$) & \phantom{1}93.1 ($\pm 0.6$) & \phantom{1}93.7 ($\pm 0.4$) \\
RefCOCO (testA)~\cite{DBLP:conf/eccv/YuPYBB16-refcoco} & \phantom{1}75.7 ($\pm 0.2$) & \phantom{1}77.2 ($\pm 0.1$) & \phantom{1}76.8 ($\pm 0.1$) & \phantom{1}78.6 ($\pm 0.3$) & \phantom{1}79.7 ($\pm 0.1$) & \phantom{1}79.3 ($\pm 0.1$) \\
RefCOCO (testB)~\cite{DBLP:conf/eccv/YuPYBB16-refcoco} & \phantom{1}71.0 ($\pm 0.3$) & \phantom{1}74.2 ($\pm 0.3$) & \phantom{1}73.9 ($\pm 0.1$) & \phantom{1}73.5 ($\pm 0.1$) & \phantom{1}76.2 ($\pm 0.3$) & \phantom{1}74.8 ($\pm 0.1$) \\
RefCOCO (val)~\cite{DBLP:conf/eccv/YuPYBB16-refcoco} & \phantom{1}73.4 ($\pm 0.1$) & \phantom{1}75.9 ($\pm 0.1$) & \phantom{1}75.0 ($\pm 0.0$) & \phantom{1}76.3 ($\pm 0.1$) & \phantom{1}78.2 ($\pm 0.1$) & \phantom{1}77.3 ($\pm 0.1$) \\
RefCOCO+ (testA)~\cite{kazemzadeh2014referit} & \phantom{1}72.7 ($\pm 0.2$) & \phantom{1}74.7 ($\pm 0.2$) & \phantom{1}73.6 ($\pm 0.2$) & \phantom{1}76.1 ($\pm 0.2$) & \phantom{1}77.7 ($\pm 0.2$) & \phantom{1}76.6 ($\pm 0.1$) \\
RefCOCO+ (testB)~\cite{kazemzadeh2014referit} & \phantom{1}64.2 ($\pm 0.2$) & \phantom{1}68.4 ($\pm 0.3$) & \phantom{1}67.1 ($\pm 0.1$) & \phantom{1}67.0 ($\pm 0.3$) & \phantom{1}71.1 ($\pm 0.2$) & \phantom{1}68.6 ($\pm 0.1$) \\
RefCOCO+ (val)~\cite{kazemzadeh2014referit} & \phantom{1}68.6 ($\pm 0.1$) & \phantom{1}72.0 ($\pm 0.2$) & \phantom{1}70.3 ($\pm 0.2$) & \phantom{1}72.1 ($\pm 0.3$) & \phantom{1}74.4 ($\pm 0.1$) & \phantom{1}72.8 ($\pm 0.1$) \\
RefCOCOg (test)~\cite{refcocog} & \phantom{1}69.0 ($\pm 0.2$) & \phantom{1}71.9 ($\pm 0.1$) & \phantom{1}70.7 ($\pm 0.1$) & \phantom{1}72.7 ($\pm 0.1$) & \phantom{1}74.8 ($\pm 0.1$) & \phantom{1}73.7 ($\pm 0.1$) \\
RefCOCOg (val)~\cite{refcocog} & \phantom{1}68.3 ($\pm 0.3$) & \phantom{1}71.4 ($\pm 0.2$) & \phantom{1}70.5 ($\pm 0.1$) & \phantom{1}72.3 ($\pm 0.2$) & \phantom{1}74.4 ($\pm 0.1$) & \phantom{1}73.0 ($\pm 0.1$) \\
ST-VQA (val)~\cite{Biten_2019-STVQA} & \phantom{1}61.9 ($\pm 0.1$) & \phantom{1}64.3 ($\pm 0.4$) & \phantom{1}65.1 ($\pm 0.4$) & \phantom{1}80.5 ($\pm 0.1$) & \phantom{1}82.0 ($\pm 0.3$) & \phantom{1}81.8 ($\pm 0.1$) \\
SciCap~\cite{hsu2021scicap} & 165.1 ($\pm 0.5$) & 159.5 ($\pm 0.7$) & 156.9 ($\pm 1.0$) & 183.3 ($\pm 0.7$) & 177.2 ($\pm 0.3$) & 172.7 ($\pm 1.5$) \\
ScienceQA~\cite{lu2022learn-scienceqa} & \phantom{1}96.1 ($\pm 0.3$) & \phantom{1}98.2 ($\pm 0.2$) & \phantom{1}98.2 ($\pm 0.2$) & \phantom{1}96.2 ($\pm 0.2$) & \phantom{1}98.5 ($\pm 0.2$) & \phantom{1}98.6 ($\pm 0.2$) \\
Screen2Words~\cite{wang2021screen2words} & 113.3 ($\pm 0.8$) & 117.8 ($\pm 0.7$) & 122.8 ($\pm 0.5$) & 114.0 ($\pm 0.5$) & 119.1 ($\pm 1.9$) & 123.4 ($\pm 0.8$) \\
TallyQA (complex)~\cite{acharya2019tallyqa} & \phantom{1}70.3 ($\pm 0.3$) & \phantom{1}73.4 ($\pm 0.1$) & \phantom{1}74.2 ($\pm 0.1$) & \phantom{1}73.6 ($\pm 0.2$) & \phantom{1}76.7 ($\pm 0.3$) & \phantom{1}76.8 ($\pm 0.2$) \\
TallyQA (simple)~\cite{acharya2019tallyqa} & \phantom{1}81.8 ($\pm 0.1$) & \phantom{1}83.2 ($\pm 0.1$) & \phantom{1}83.4 ($\pm 0.1$) & \phantom{1}85.3 ($\pm 0.1$) & \phantom{1}86.2 ($\pm 0.1$) & \phantom{1}85.7 ($\pm 0.1$) \\
TextCaps~\cite{sidorov2019textcaps} & 127.5 ($\pm 0.3$) & 137.9 ($\pm 0.3$) & 139.9 ($\pm 0.4$) & 152.1 ($\pm 0.3$) & 157.7 ($\pm 0.7$) & 153.6 ($\pm 0.5$) \\
TextVQA (val)~\cite{singh2019towards-textvqa} & \phantom{1}59.6 ($\pm 0.3$) & \phantom{1}64.0 ($\pm 0.3$) & \phantom{1}64.7 ($\pm 0.2$) & \phantom{1}75.2 ($\pm 0.2$) & \phantom{1}76.6 ($\pm 0.1$) & \phantom{1}76.2 ($\pm 0.1$) \\
VATEX~\cite{wang2019vatex} & \phantom{1}80.8 ($\pm 0.4$) & \phantom{1}82.7 ($\pm 0.5$) & \phantom{100}-\phantom{0 ($\pm 0.0$)} & \phantom{100}-\phantom{0 ($\pm 0.0$)} & \phantom{100}-\phantom{0 ($\pm 0.0$)} & \phantom{100}-\phantom{0 ($\pm 0.0$)} \\
VQAv2 (minival)~\cite{balanced_vqa_v2} & \phantom{1}83.0 ($\pm 0.2$) & \phantom{1}84.3 ($\pm 0.2$) & \phantom{1}84.5 ($\pm 0.1$) & \phantom{1}84.8 ($\pm 0.2$) & \phantom{1}85.8 ($\pm 0.1$) & \phantom{1}85.8 ($\pm 0.2$) \\
VizWizVQA (val)~\cite{gurari2018vizwiz} & \phantom{1}76.4 ($\pm 0.4$) & \phantom{1}78.1 ($\pm 0.4$) & \phantom{1}78.7 ($\pm 0.2$) & \phantom{1}77.5 ($\pm 0.2$) & \phantom{1}78.6 ($\pm 0.4$) & \phantom{1}78.9 ($\pm 0.5$) \\
WidgetCap~\cite{Li2020WidgetCap} & 138.1 ($\pm 0.7$) & 139.8 ($\pm 1.0$) & 138.8 ($\pm 0.8$) & 151.4 ($\pm 0.8$) & 151.9 ($\pm 0.4$) & 148.9 ($\pm 0.7$) \\
XM3600 (avg35)~\cite{thapliyal-etal-2022-crossmodal-COCO35L-XM3600} & \phantom{1}42.8 ($\pm 0.1$) & \phantom{1}44.5 ($\pm 0.1$) & \phantom{1}45.2 ($\pm 0.1$) & \phantom{1}43.2 ($\pm 0.1$) & \phantom{1}44.6 ($\pm 0.1$) & \phantom{1}45.2 ($\pm 0.1$) \\
XM3600 (en)~\cite{thapliyal-etal-2022-crossmodal-COCO35L-XM3600} & \phantom{1}79.8 ($\pm 0.7$) & \phantom{1}80.7 ($\pm 0.3$) & \phantom{1}81.0 ($\pm 0.9$) & \phantom{1}80.3 ($\pm 0.8$) & \phantom{1}81.5 ($\pm 0.4$) & \phantom{1}81.0 ($\pm 0.2$) \\
xGQA (avg7)~\cite{pfeiffer-etal-2022-xgqa} & \phantom{1}58.6 ($\pm 0.2$) & \phantom{1}61.4 ($\pm 0.1$) & \phantom{1}61.1 ($\pm 0.1$) & \phantom{1}60.4 ($\pm 0.2$) & \phantom{1}62.6 ($\pm 0.2$) & \phantom{1}62.1 ($\pm 0.3$) \\
\bottomrule
\end{tabular}

  }
  \caption{Mean and std-deviation over 5 finetuning runs of PaliGemma 3B, 10B, 28B models at 224px$^2$ and 448px$^2$ resolutions on over 30+ academic tasks from \cite{beyer2024paligemma}. Tasks splits, preprocessing, metrics and hyper-parameters following the 224px$^2$ versions according to previous work. Only the learning rate has been selected per model size based on validation splits.
  }
  \label{tab:app:pg-results}
\end{table}

{\small
\newcommand{\con}{\rowcolor{gray!15}}

\begin{longtable}{lrrrrrrrr}

\caption{Sweep of learning rates on the various tasks and model sizes at 224px$^2$ resolution. Although we report numbers in all metrics, learning rate selection was done based on the validation split and not on the zero-shot numbers.} \label{tab:app:pg_lr_sweep} \\
\toprule
 &  & 3e-7 & 6e-7 & 1e-6 & 3e-6 & 6e-6 & 1e-5 & 3e-5 \\
Task & Model &  &  &  &  &  &  &  \\
\midrule
\endfirsthead

\caption[]{Sweep of learning rates on the various tasks and model sizes at 224px$^2$ resolution. Although we report numbers in all metrics, learning rate selection was done based on the validation split and not on the zero-shot numbers.} \\
\toprule
 &  & 3e-7 & 6e-7 & 1e-6 & 3e-6 & 6e-6 & 1e-5 & 3e-5 \\
Task & Model &  &  &  &  &  &  &  \\
\midrule
\endhead

\midrule
\multicolumn{9}{r}{Continued on next page} \\
\midrule
\endfoot
\bottomrule
\endlastfoot

\con & \texttt{3B} & 61.8 & 67.6 & 70.6 & 75.0 & 76.9 & 75.1 & 68.8 \\
\con AI2D (minival) & \texttt{10B} & 80.0 & 82.9 & 85.3 & 84.4 & 82.9 & 82.1 & 69.2 \\
\con & \texttt{28B} & 81.9 & 82.3 & 83.2 & 85.9 & 85.0 & 83.4 & 75.7 \\
\multirow[c]{3}{*}{AOKVQA-DA (val)} & \texttt{3B} & 59.3 & 62.9 & 64.0 & 64.6 & 63.6 & 59.3 & 52.8 \\
 & \texttt{10B} & 67.7 & 68.6 & 68.8 & 66.6 & 64.6 & 57.3 & 50.5 \\
 & \texttt{28B} & 69.7 & 70.2 & 69.8 & 69.0 & 66.3 & 60.8 & 51.1 \\
\con & \texttt{3B} & 76.9 & 78.7 & 79.4 & 80.8 & 77.2 & 76.9 & 63.8 \\
\con AOKVQA-MC (val) & \texttt{10B} & 83.8 & 83.3 & 83.3 & 82.7 & 79.4 & 75.5 & 56.1 \\
\con & \texttt{28B} & 83.3 & 84.0 & 85.1 & 82.5 & 82.4 & 78.2 & 58.4 \\
\multirow[c]{2}{*}{ActivityNet-CAP (minival)} & \texttt{3B} & 26.1 & 28.5 & 28.5 & 30.6 & 30.0 & 30.6 & 29.8 \\
 & \texttt{10B} & 28.6 & 31.4 & 30.8 & 31.6 & 30.0 & 31.1 & 28.6 \\
\con ActivityNet-QA (minival) & \texttt{3B} & 43.3 & 46.8 & 49.4 & 52.6 & 53.8 & 53.5 & 52.0 \\
\con & \texttt{10B} & 49.9 & 52.2 & 53.9 & 55.0 & 55.3 & 54.6 & 51.2 \\
\multirow[c]{3}{*}{COCO-35L (avg34)} & \texttt{3B} & 110.1 & 111.8 & 113.6 & 113.9 & 113.6 & 113.2 & 111.7 \\
 & \texttt{10B} & 115.4 & 115.8 & 115.2 & 113.6 & 112.9 & 112.2 & 111.7 \\
 & \texttt{28B} & 116.7 & 116.6 & 115.4 & 114.0 & 112.1 & 111.2 & 109.6 \\
\con & \texttt{3B} & 137.9 & 138.6 & 139.1 & 138.4 & 137.6 & 136.5 & 133.8 \\
\con COCO-35L (en) & \texttt{10B} & 140.6 & 140.3 & 139.6 & 137.3 & 135.5 & 133.8 & 132.5 \\
\con & \texttt{28B} & 142.5 & 141.3 & 140.4 & 137.7 & 134.5 & 133.2 & 129.9 \\
\multirow[c]{3}{*}{COCOcap (minival)} & \texttt{3B} & 146.3 & 146.7 & 145.4 & 147.2 & 147.1 & 147.0 & 142.0 \\
 & \texttt{10B} & 148.3 & 149.4 & 148.2 & 148.3 & 147.0 & 146.5 & 143.6 \\
 & \texttt{28B} & 148.8 & 149.5 & 149.2 & 149.5 & 148.2 & 145.3 & 145.7 \\
\con & \texttt{3B} & 60.8 & 64.3 & 66.0 & 69.7 & 69.5 & 68.4 & 63.6 \\
\con ChartQA (aug) (minival) & \texttt{10B} & 69.0 & 68.6 & 71.1 & 69.5 & 69.9 & 68.4 & 60.4 \\
\con  & \texttt{28B} & 66.8 & 63.4 & 65.2 & 66.7 & 66.0 & 64.1 & 55.9 \\
\multirow[c]{3}{*}{ChartQA (human) (minival)} & \texttt{3B} & 41.4 & 42.8 & 42.7 & 44.1 & 43.2 & 42.9 & 35.4 \\
 & \texttt{10B} & 50.9 & 50.8 & 50.8 & 49.2 & 47.0 & 44.5 & 34.6 \\
 & \texttt{28B} & 48.3 & 46.9 & 47.7 & 46.5 & 45.3 & 41.8 & 33.8 \\
\con & \texttt{3B} & 82.7 & 82.9 & 82.0 & 79.0 & 82.0 & 78.0 & 70.4 \\
\con CountBenchQA & \texttt{10B} & 88.2 & 84.7 & 85.1 & 82.9 & 81.4 & 78.2 & 65.7 \\
\con & \texttt{28B} & 87.8 & 88.4 & 88.4 & 88.6 & 86.7 & 83.3 & 69.6 \\
\multirow[c]{3}{*}{DocVQA (val)} & \texttt{3B} & 37.8 & 37.9 & 37.3 & 39.4 & 40.2 & 38.7 & 32.5 \\
 & \texttt{10B} & 42.4 & 40.9 & 42.2 & 44.1 & 41.4 & 39.8 & 29.6 \\
 & \texttt{28B} & 42.7 & 42.1 & 43.1 & 45.2 & 42.1 & 40.5 & 30.9 \\
\con & \texttt{3B} & 70.9 & 72.2 & 72.9 & 73.9 & 73.9 & 73.8 & 72.4 \\
\con GQA (minival) & \texttt{10B} & 73.6 & 74.3 & 74.7 & 74.4 & 74.4 & 74.2 & 71.5 \\
\con & \texttt{28B} & 73.7 & 73.9 & 74.7 & 74.8 & 74.6 & 74.1 & 72.3 \\
\multirow[c]{3}{*}{InfoVQA (val)} & \texttt{3B} & 21.6 & 22.9 & 23.8 & 25.4 & 25.2 & 25.1 & 22.3 \\
 & \texttt{10B} & 33.4 & 33.5 & 33.2 & 33.2 & 32.2 & 29.8 & 21.7 \\
 & \texttt{28B} & 36.9 & 36.6 & 36.3 & 36.2 & 35.5 & 34.1 & 25.4 \\
\con & \texttt{3B} & 69.9 & 73.4 & 77.1 & 81.2 & 83.0 & 82.4 & 69.9 \\
\con MARVL (avg5) & \texttt{10B} & 86.5 & 88.2 & 89.2 & 89.4 & 89.1 & 87.4 & 67.6 \\
\con & \texttt{28B} & 86.7 & 88.5 & 89.5 & 90.3 & 90.8 & 89.2 & 76.2 \\
\multirow[c]{2}{*}{MSRVTT-CAP (minival)} & \texttt{3B} & 62.8 & 66.1 & 67.8 & 67.6 & 72.6 & 74.0 & 68.3 \\
 & \texttt{10B} & 70.4 & 71.5 & 75.3 & 74.0 & 66.2 & 69.4 & 67.2 \\
\con MSRVTT-QA (minival) & \texttt{3B} & 44.1 & 47.0 & 48.5 & 51.1 & 52.0 & 51.2 & 49.9 \\
\con & \texttt{10B} & 49.3 & 51.2 & 51.9 & 53.2 & 53.1 & 52.1 & 49.7 \\
\multirow[c]{2}{*}{MSVD-QA (minival)} & \texttt{3B} & 55.2 & 57.8 & 60.7 & 63.3 & 63.1 & 61.3 & 57.0 \\
 & \texttt{10B} & 61.1 & 63.9 & 65.4 & 64.2 & 63.2 & 63.0 & 56.3 \\
\con & \texttt{3B} & 82.5 & 86.2 & 88.2 & 90.4 & 90.9 & 90.2 & 85.9 \\
\con NLVR2 (minival) & \texttt{10B} & 91.8 & 93.0 & 93.3 & 93.3 & 92.5 & 91.7 & 86.1 \\
\con & \texttt{28B} & 92.2 & 92.8 & 93.6 & 93.7 & 93.7 & 92.2 & 88.0 \\
\multirow[c]{3}{*}{NoCaps } & \texttt{3B} & 123.3 & 123.6 & 124.0 & 123.4 & 122.5 & 120.5 & 112.3 \\
 & \texttt{10B} & 126.7 & 126.1 & 126.0 & 125.2 & 122.1 & 120.5 & 111.5 \\
 & \texttt{28B} & 127.5 & 127.5 & 126.5 & 124.0 & 123.0 & 120.3 & 113.0 \\
\con & \texttt{3B} & 72.6 & 73.1 & 73.4 & 73.4 & 73.2 & 72.9 & 70.6 \\
\con OCR-VQA (minival) & \texttt{10B} & 74.7 & 74.5 & 74.3 & 73.9 & 73.5 & 73.0 & 70.6 \\
\con & \texttt{28B} & 75.5 & 75.5 & 75.2 & 74.8 & 73.9 & 72.5 & 71.0 \\
\multirow[c]{3}{*}{OKVQA (minival)} & \texttt{3B} & 49.4 & 52.3 & 54.3 & 57.6 & 56.2 & 52.9 & 47.2 \\
 & \texttt{10B} & 57.8 & 60.5 & 61.3 & 60.8 & 58.7 & 55.6 & 44.1 \\
 & \texttt{28B} & 64.6 & 64.4 & 65.4 & 63.8 & 60.6 & 56.8 & 46.4 \\
\con & \texttt{3B} & 92.8 & 93.2 & 93.3 & 93.0 & 93.3 & 93.4 & 93.3 \\
\con RSVQA-hr (minival) & \texttt{10B} & 93.3 & 93.2 & 93.1 & 93.0 & 93.4 & 93.3 & 89.4 \\
\con & \texttt{28B} & 93.1 & 93.4 & 93.3 & 93.3 & 93.3 & 93.3 & 92.9 \\
\multirow[c]{3}{*}{RSVQA-lr (minival)} & \texttt{3B} & 90.7 & 92.4 & 92.7 & 93.3 & 92.1 & 92.2 & 92.3 \\
 & \texttt{10B} & 92.3 & 92.7 & 92.0 & 91.7 & 91.8 & 92.8 & 92.0 \\
 & \texttt{28B} & 91.8 & 92.1 & 92.4 & 92.7 & 92.9 & 92.9 & 92.3 \\
\con & \texttt{3B} & 73.1 & 74.5 & 75.3 & 75.5 & 75.8 & 75.8 & 74.1 \\
\con RefCOCO (testA) & \texttt{10B} & 76.7 & 76.9 & 77.1 & 77.2 & 77.1 & 76.1 & 71.6 \\
\con & \texttt{28B} & 76.2 & 76.7 & 76.8 & 76.8 & 76.6 & 75.5 & 71.6 \\
\multirow[c]{3}{*}{RefCOCO (testB)} & \texttt{3B} & 68.0 & 70.1 & 70.8 & 71.2 & 70.8 & 70.9 & 69.7 \\
 & \texttt{10B} & 73.8 & 74.3 & 74.3 & 74.2 & 73.4 & 73.4 & 68.6 \\
 & \texttt{28B} & 73.0 & 73.9 & 73.8 & 72.8 & 73.1 & 72.0 & 68.4 \\
\con & \texttt{3B} & 70.4 & 72.1 & 73.0 & 73.2 & 73.3 & 73.4 & 71.6 \\
\con RefCOCO (val) & \texttt{10B} & 75.1 & 75.6 & 75.8 & 76.1 & 75.6 & 74.9 & 70.6 \\
\con & \texttt{28B} & 74.6 & 75.0 & 75.2 & 74.8 & 74.6 & 74.0 & 69.9 \\
\multirow[c]{3}{*}{RefCOCO+ (testA)} & \texttt{3B} & 67.6 & 70.1 & 70.8 & 71.8 & 72.2 & 72.7 & 71.0 \\
 & \texttt{10B} & 72.9 & 73.5 & 74.0 & 75.0 & 74.9 & 74.2 & 69.0 \\
 & \texttt{28B} & 72.7 & 73.4 & 73.4 & 74.0 & 74.3 & 72.9 & 69.3 \\
\con & \texttt{3B} & 55.3 & 58.6 & 60.5 & 62.9 & 63.2 & 64.6 & 63.8 \\
\con RefCOCO+ (testB) & \texttt{10B} & 66.0 & 67.1 & 67.3 & 68.4 & 68.2 & 67.9 & 62.6 \\
\con & \texttt{28B} & 65.3 & 66.4 & 67.1 & 67.5 & 67.8 & 67.0 & 62.7 \\
\multirow[c]{3}{*}{RefCOCO+ (val)} & \texttt{3B} & 61.3 & 64.2 & 65.8 & 67.0 & 67.9 & 68.6 & 67.5 \\
 & \texttt{10B} & 69.8 & 70.8 & 71.1 & 72.0 & 71.8 & 71.3 & 66.5 \\
 & \texttt{28B} & 69.0 & 70.0 & 70.4 & 70.8 & 71.0 & 70.4 & 65.7 \\
\con & \texttt{3B} & 65.5 & 67.2 & 68.4 & 68.7 & 68.9 & 69.0 & 67.2 \\
\con RefCOCOg (test) & \texttt{10B} & 70.9 & 71.6 & 71.6 & 71.7 & 71.3 & 70.4 & 65.2 \\
\con & \texttt{28B} & 69.9 & 70.5 & 70.8 & 70.7 & 70.6 & 69.7 & 64.9 \\
\multirow[c]{3}{*}{RefCOCOg (val)} & \texttt{3B} & 65.2 & 67.0 & 67.8 & 68.0 & 68.0 & 68.2 & 66.1 \\
 & \texttt{10B} & 70.8 & 71.4 & 71.4 & 71.4 & 71.0 & 70.0 & 64.9 \\
 & \texttt{28B} & 69.9 & 70.4 & 70.2 & 70.2 & 70.1 & 69.2 & 64.0 \\
\con & \texttt{3B} & 56.1 & 58.8 & 60.4 & 61.5 & 62.3 & 61.2 & 57.0 \\
\con ST-VQA (val) & \texttt{10B} & 60.9 & 62.9 & 63.8 & 64.0 & 63.9 & 61.2 & 54.8 \\
\con & \texttt{28B} & 63.0 & 64.4 & 65.2 & 65.5 & 64.3 & 62.6 & 55.7 \\
\multirow[c]{3}{*}{SciCap (minival)} & \texttt{3B} & 55.2 & 67.4 & 76.9 & 109.4 & 130.3 & 138.8 & 148.1 \\
 & \texttt{10B} & 78.6 & 92.5 & 106.2 & 128.1 & 136.9 & 143.2 & 143.8 \\
 & \texttt{28B} & 80.3 & 94.7 & 104.0 & 125.9 & 136.2 & 140.1 & 141.7 \\
\con & \texttt{3B} & 87.7 & 92.1 & 94.5 & 95.1 & 95.2 & 94.3 & 91.4 \\
\con ScienceQA (minival) & \texttt{10B} & 96.9 & 97.1 & 97.6 & 97.6 & 97.1 & 96.2 & 93.7 \\
\con & \texttt{28B} & 96.8 & 97.1 & 97.4 & 97.2 & 96.8 & 96.1 & 94.2 \\
\multirow[c]{3}{*}{Screen2Words (minival)} & \texttt{3B} & 95.1 & 104.2 & 109.0 & 109.3 & 113.2 & 112.5 & 110.1 \\
 & \texttt{10B} & 110.9 & 115.4 & 118.2 & 118.1 & 114.7 & 113.0 & 110.0 \\
 & \texttt{28B} & 113.0 & 119.5 & 120.4 & 118.8 & 116.2 & 114.2 & 106.3 \\
\con & \texttt{3B} & 66.6 & 67.8 & 68.6 & 70.0 & 70.0 & 70.5 & 66.7 \\
\con TallyQA (complex) & \texttt{10B} & 72.0 & 72.5 & 73.4 & 73.5 & 72.7 & 72.0 & 65.8 \\
\con & \texttt{28B} & 73.1 & 73.5 & 73.9 & 74.8 & 73.8 & 73.0 & 68.1 \\
\multirow[c]{3}{*}{TallyQA (simple)} & \texttt{3B} & 80.4 & 81.1 & 81.3 & 81.8 & 81.9 & 81.5 & 79.1 \\
 & \texttt{10B} & 83.0 & 83.3 & 83.1 & 83.2 & 82.7 & 82.1 & 79.1 \\
 & \texttt{28B} & 82.9 & 83.3 & 83.3 & 83.5 & 83.0 & 82.2 & 79.7 \\
\con & \texttt{3B} & 122.8 & 131.9 & 136.5 & 136.2 & 133.6 & 132.8 & 126.0 \\
\con TextCaps (minival) & \texttt{10B} & 140.3 & 145.3 & 145.4 & 145.4 & 144.2 & 141.0 & 125.8 \\
\con & \texttt{28B} & 150.9 & 149.0 & 150.2 & 145.5 & 144.0 & 142.1 & 126.2 \\
\multirow[c]{3}{*}{TextVQA (val)} & \texttt{3B} & 57.6 & 58.7 & 59.3 & 59.6 & 59.4 & 58.0 & 51.1 \\
 & \texttt{10B} & 63.4 & 64.1 & 63.9 & 63.2 & 61.6 & 58.1 & 48.3 \\
 & \texttt{28B} & 64.5 & 64.7 & 65.3 & 64.8 & 63.3 & 59.3 & 49.9 \\
\con VATEX (minival) & \texttt{3B} & 84.4 & 87.2 & 89.8 & 90.7 & 90.2 & 90.2 & 86.3 \\
\con & \texttt{10B} & 91.4 & 93.2 & 93.4 & 93.7 & 90.4 & 89.9 & 84.5 \\
\multirow[c]{3}{*}{} & \texttt{3B} & 80.9 & 81.5 & 82.1 & 82.7 & 82.4 & 81.9 & 79.6 \\
 & \texttt{10B} & 83.8 & 84.1 & 84.3 & 83.7 & 83.1 & 82.0 & 79.4 \\
 & \texttt{28B} & 83.8 & 84.1 & 84.1 & 83.8 & 82.8 & 82.0 & 79.7 \\
\con & \texttt{3B} & 72.5 & 74.2 & 74.8 & 76.4 & 76.6 & 76.7 & 74.0 \\
\con VizWizVQA (val) & \texttt{10B} & 76.1 & 77.1 & 77.8 & 78.0 & 77.3 & 77.2 & 73.3 \\
\con & \texttt{28B} & 76.3 & 77.6 & 78.2 & 78.8 & 77.8 & 76.7 & 72.5 \\
\multirow[c]{3}{*}{WidgetCap (minival)} & \texttt{3B} & 137.0 & 141.9 & 141.8 & 142.3 & 141.7 & 140.6 & 129.7 \\
 & \texttt{10B} & 146.3 & 148.4 & 150.9 & 148.2 & 144.5 & 140.8 & 133.3 \\
 & \texttt{28B} & 144.0 & 147.6 & 145.9 & 147.0 & 144.1 & 143.0 & 133.0 \\
\con & \texttt{3B} & 44.2 & 43.9 & 43.7 & 42.7 & 41.7 & 40.8 & 37.8 \\
\con XM3600 (avg35) & \texttt{10B} & 45.0 & 44.5 & 43.9 & 42.1 & 40.7 & 39.3 & 36.8 \\
\con & \texttt{28B} & 45.2 & 44.6 & 44.0 & 42.3 & 41.1 & 39.1 & 35.8 \\
\multirow[c]{3}{*}{} & \texttt{3B} & 83.7 & 83.1 & 82.2 & 79.1 & 78.3 & 76.9 & 70.9 \\
 & \texttt{10B} & 82.5 & 80.6 & 78.6 & 75.0 & 73.0 & 72.0 & 69.9 \\
 & \texttt{28B} & 80.9 & 79.8 & 79.4 & 76.4 & 73.6 & 71.3 & 66.1 \\
\con & \texttt{3B} & 51.7 & 54.0 & 55.3 & 58.0 & 58.7 & 57.8 & 49.1 \\
\con xGQA (avg7) & \texttt{10B} & 58.5 & 60.5 & 61.4 & 61.3 & 61.8 & 60.2 & 38.0 \\
\con & \texttt{28B} & 58.8 & 59.2 & 60.8 & 62.3 & 61.9 & 61.7 & 49.4 \\
\end{longtable}

}

\newcommand{\gain}[1]{{\textcolor{gray}{(#1)}}}
\newcommand{\loss}[1]{{\textcolor{gray}{(#1)}}}
\newcommand{\zsrow}{$\llcorner$}

\begin{table}
\centering
\small
\begin{tabular}{lrrrr}
\toprule
& \multicolumn{2}{c}{224px$^2$} & \multicolumn{2}{c}{448px$^2$} \\
Task & \multicolumn{1}{c}{PG1} & \multicolumn{1}{c}{PG2} & \multicolumn{1}{c}{PG1} & \multicolumn{1}{c}{PG2} \\
\midrule
AI2D & 72.1 & 74.7 \gain{$+2.6$} & 73.3 & 76.0 \gain{$+2.7$} \\
AOKVQA-DA (val) & 61.1 & 64.2 \gain{$+3.1$} & 65.7 & 67.9 \gain{$+2.2$} \\
AOKVQA-MC (val) & 78.5 & 79.7 \gain{$+1.2$} & 80.3 & 82.5 \gain{$+2.2$} \\
ActivityNet-CAP & 34.6 & 34.2 \loss{$-0.4$} & -\phantom{0} & -\phantom{0 \gain{$+0.0$}} \\
ActivityNet-QA & 50.8 & 51.3 \gain{$+0.5$} & -\phantom{0} & -\phantom{0 \gain{$+0.0$}} \\
COCO-35L (avg34) & 113.7 & 113.9 \gain{$+0.2$} & 115.8 & 115.8 \gain{$+0.0$} \\
COCO-35L (en) & 139.2 & 138.4 \loss{$-0.8$} & 141.2 & 140.4 \loss{$-0.8$} \\
COCOcap & 141.9 & 141.3 \loss{$-0.6$} & 144.6 & 143.4 \loss{$-1.2$} \\
ChartQA (aug) & 74.2 & 74.4 \gain{$+0.2$} & 88.5 & 89.2 \gain{$+0.7$} \\
ChartQA (human) & 40.0 & 42.0 \gain{$+2.0$} & 54.2 & 54.0 \loss{$-0.2$} \\
CountBenchQA & 81.9 & 81.0 \loss{$-0.9$} & 83.1 & 82.0 \loss{$-1.1$} \\
DocVQA (val) & 37.8 & 39.9 \gain{$+2.1$} & 74.1 & 73.6 \loss{$-0.5$} \\
GQA & 65.6 & 66.2 \gain{$+0.6$} & 67.0 & 68.1 \gain{$+1.1$} \\
InfoVQA (val) & 25.5 & 25.2 \loss{$-0.3$} & 37.0 & 37.5 \gain{$+0.5$} \\
MARVL (avg5) & 80.6 & 83.5 \gain{$+2.9$} & 76.8 & 82.7 \gain{$+5.9$} \\
MSRVTT-CAP & 70.5 & 68.5 \loss{$-2.0$} & -\phantom{0} & -\phantom{0 \gain{$+0.0$}} \\
MSRVTT-QA & 50.1 & 50.5 \gain{$+0.4$} & -\phantom{0} & -\phantom{0 \gain{$+0.0$}} \\
MSVD-QA & 60.2 & 61.1 \gain{$+0.9$} & -\phantom{0} & -\phantom{0 \gain{$+0.0$}} \\
NLVR2 & 90.0 & 91.4 \gain{$+1.4$} & 88.9 & 91.6 \gain{$+2.7$} \\
NoCaps & 121.7 & 123.1 \gain{$+1.4$} & 123.6 & 123.5 \loss{$-0.1$} \\
OCR-VQA & 72.3 & 73.4 \gain{$+1.1$} & 74.6 & 75.7 \gain{$+1.1$} \\
OKVQA & 63.5 & 64.2 \gain{$+0.7$} & 63.2 & 64.1 \gain{$+0.9$} \\
RSVQA-hr (test) & 92.6 & 92.7 \gain{$+0.1$} & 92.8 & 92.8 \gain{$+0.0$} \\
RSVQA-hr (test2) & 90.6 & 90.9 \gain{$+0.3$} & 90.5 & 90.7 \gain{$+0.2$} \\
RSVQA-lr & 92.6 & 93.0 \gain{$+0.4$} & 93.1 & 92.7 \loss{$-0.4$} \\
RefCOCO (testA) & 75.7 & 75.7 \gain{$+0.0$} & 77.9 & 78.6 \gain{$+0.7$} \\
RefCOCO (testB) & 70.7 & 71.0 \gain{$+0.3$} & 72.4 & 73.5 \gain{$+1.1$} \\
RefCOCO (val) & 73.4 & 73.4 \gain{$+0.0$} & 75.6 & 76.3 \gain{$+0.7$} \\
RefCOCO+ (testA) & 71.9 & 72.7 \gain{$+0.8$} & 74.2 & 76.1 \gain{$+1.9$} \\
RefCOCO+ (testB) & 64.5 & 64.2 \loss{$-0.3$} & 64.5 & 67.0 \gain{$+2.5$} \\
RefCOCO+ (val) & 68.3 & 68.6 \gain{$+0.3$} & 69.8 & 72.1 \gain{$+2.3$} \\
RefCOCOg (test) & 68.2 & 69.0 \gain{$+0.8$} & 71.0 & 72.7 \gain{$+1.7$} \\
RefCOCOg (val) & 67.7 & 68.3 \gain{$+0.6$} & 70.1 & 72.3 \gain{$+2.2$} \\
ST-VQA (val) & 61.6 & 61.9 \gain{$+0.3$} & 79.7 & 80.5 \gain{$+0.8$} \\
SciCap & 162.3 & 165.1 \gain{$+2.8$} & 181.5 & 183.3 \gain{$+1.8$} \\
ScienceQA & 95.4 & 96.1 \gain{$+0.7$} & 95.9 & 96.2 \gain{$+0.3$} \\
Screen2Words & 117.6 & 113.3 \loss{$-4.3$} & 119.6 & 114.0 \loss{$-5.6$} \\
TallyQA (complex) & 69.6 & 70.3 \gain{$+0.7$} & 72.3 & 73.6 \gain{$+1.3$} \\
TallyQA (simple) & 81.7 & 81.8 \gain{$+0.1$} & 84.9 & 85.3 \gain{$+0.4$} \\
TextCaps & 127.5 & 127.5 \gain{$+0.0$} & 153.9 & 152.1 \loss{$-1.8$} \\
TextVQA (val) & 59.0 & 59.6 \gain{$+0.6$} & 74.6 & 75.2 \gain{$+0.6$} \\
VATEX & 79.7 & 80.8 \gain{$+1.1$} & -\phantom{0} & -\phantom{0 \gain{$+0.0$}} \\
VQAv2 (minival) & 82.1 & 83.0 \gain{$+0.9$} & 84.6 & 84.8 \gain{$+0.2$} \\
VizWizVQA (val) & 73.7 & 76.4 \gain{$+2.7$} & 75.5 & 77.5 \gain{$+2.0$} \\
WidgetCap & 136.1 & 138.1 \gain{$+2.0$} & 148.4 & 151.4 \gain{$+3.0$} \\
XM3600 (avg35) & 41.9 & 42.8 \gain{$+0.9$} & 42.4 & 43.2 \gain{$+0.8$} \\
XM3600 (en) & 78.0 & 79.8 \gain{$+1.8$} & 80.0 & 80.3 \gain{$+0.3$} \\
xGQA (avg7) & 57.3 & 58.6 \gain{$+1.3$} & 57.9 & 60.4 \gain{$+2.5$} \\
\bottomrule

\end{tabular}
\caption{Comparison of \pg 3B and \pgtwo 3B at 224px$^2$ and 448px$^2$ resolutions. PG1 and PG2 refer to \pgnospace{}~\cite{beyer2024paligemma} and \pgtwonospace{}, respectively.}
\label{tab:app:pg1_pg2}
\end{table}

\end{document}